\documentclass[journal]{IEEEtran}
\usepackage[T1]{fontenc}% optional T1 font encoding
\usepackage{graphicx}
\usepackage{times}
\usepackage{helvet}
\usepackage{courier}
\usepackage{amsmath}
\usepackage{algorithm}
\usepackage{algorithmic}
\usepackage{csquotes} 
\usepackage{color}
\usepackage{paralist}
\usepackage{amssymb}
\usepackage{indentfirst}
\usepackage{subfigure}
\usepackage{float}
\usepackage{multirow}
\usepackage{cite}
\usepackage{mathrsfs}

\usepackage{mathrsfs} 
\usepackage{amsfonts}
\usepackage{todonotes}
\usepackage{pgfplots} %引用包
\usepackage{epstopdf}
\usepackage{epsfig}
\pgfplotsset{compat=newest}
\usepackage{color}
\definecolor{forestgreen}{RGB}{0,139,69}
\usepackage{xcolor}
\definecolor{citecolor}{HTML}{0071bc}
\usepackage[colorlinks, linkcolor=red,  anchorcolor=blue, citecolor=citecolor]{hyperref} 

\usepackage{xcolor}
\definecolor{SeaGreen4}{RGB}{0,205,102} 
\definecolor{SlateBlue}{RGB}{106,90,205} 
\definecolor{DarkRed}{RGB}{178,34,34} 

\usepackage{graphicx} % Required for inserting images
\usepackage{multirow}
\usepackage{booktabs}
\usepackage[table,xcdraw]{xcolor}
\usepackage{caption}
\usepackage{array}
\usepackage[normalem]{ulem}
\useunder{\uline}{\ul}{}

\usepackage[switch]{lineno}
\usepackage{textcomp,booktabs}
\usepackage{amssymb}% http://ctan.org/pkg/amssymb
\usepackage{pifont}% http://ctan.org/pkg/pifont

\usepackage{makecell}

\usepackage{colortbl}
\definecolor{mygray}{gray}{.9}
\definecolor{mypink}{rgb}{.99,.91,.95}
\definecolor{mycyan}{cmyk}{.3,0,0,0}

\begin{document}

\title{ 
    HGTS-Former: Hierarchical HyperGraph Transformer for Multivariate Time Series Analysis 
}   

\author{Hao Si, Xiao Wang*, \emph{Member, IEEE}, Fan Zhang, Xiaoya Zhou, Dengdi Sun, Wanli Lyu, \\ 
    Qingquan Yang, Jin Tang   
\thanks{ $\bullet$ Hao Si, Xiao Wang, Fan Zhang, Xiaoya Zhou, Dengdi Sun, Wanli Lyu, Jin Tang are with the School of Computer Science and Technology, Anhui University, Hefei 230601, China. (email: \{xiaowang, tangjin\}@ahu.edu.cn)}
\thanks{ $\bullet$ Qingquan Yang is with the Institute of Plasma Physics, Chinese Academy of Sciences, Hefei, China. (email: yangqq@ipp.ac.cn)}
\thanks{* Corresponding Author: Xiao Wang (xiaowang@ahu.edu.cn)}   
}

\markboth{ IEEE Transactions on ***, 2026 } 
{Shell \MakeLowercase{\textit{et al.}}: Bare Demo of IEEEtran.cls for IEEE Journals}

% make the title area
\maketitle

\begin{abstract}
Multivariate time series analysis has long been one of the key research topics in the field of artificial intelligence. However, analyzing complex time series data remains a challenging and unresolved problem due to its high dimensionality, dynamic nature, and complex interactions among variables. 
Inspired by the strong structural modeling capability of hypergraphs, this paper proposes a novel hypergraph-based time series Transformer backbone network, termed HGTS-Former, to address the multivariate coupling in time series data. Specifically, given the multivariate time series signal, we first normalize and embed each patch into tokens. Then, we adopt the multi-head self-attention to enhance the temporal representation of each patch. The hierarchical hypergraphs are constructed to aggregate the temporal patterns within each channel and fine-grained relations between different variables. After that, we convert the hyperedge into node features through the EdgeToNode module and adopt the feed-forward network to further enhance the output features. 
Extensive experiments on multiple representative time series analysis tasks and public datasets fully validated the effectiveness of our proposed HGTS-Former. Moreover, we present EAST-ELM640, a large-scale time series dataset for Edge-Localized Mode (ELM) recognition in nuclear fusion, on which we achieve state-of-the-art performance.
The source code will be released on \url{https://github.com/Event-AHU/Time_Series_Analysis}
\end{abstract}


\IEEEpeerreviewmaketitle

\section{Introduction}

%% background 
\IEEEPARstart{M}{ultivariate} Time Series (MTS) analysis is a critical research topic in artificial intelligence and has been widely used in nuclear fusion~\cite{wang2025time}, stock prices~\cite{mehtab2020stock, jarrah2023predicting, xiang2023experimental}, and weather forecast~\cite{zheng2022multivariate, agada2023multidimensional, sanhudo2021multivariate}. These tasks are usually modeled as forecast, anomaly detection, classification, and interpolation tasks in machine learning. The core challenges of MTS analysis lie in the high dimensionality and complexity of multivariate time series data, as well as noise and missing values. As a result, there is currently no unified solution that can perfectly address all of the issues mentioned above.

\begin{figure*}
\centering
\includegraphics[width=1\linewidth]{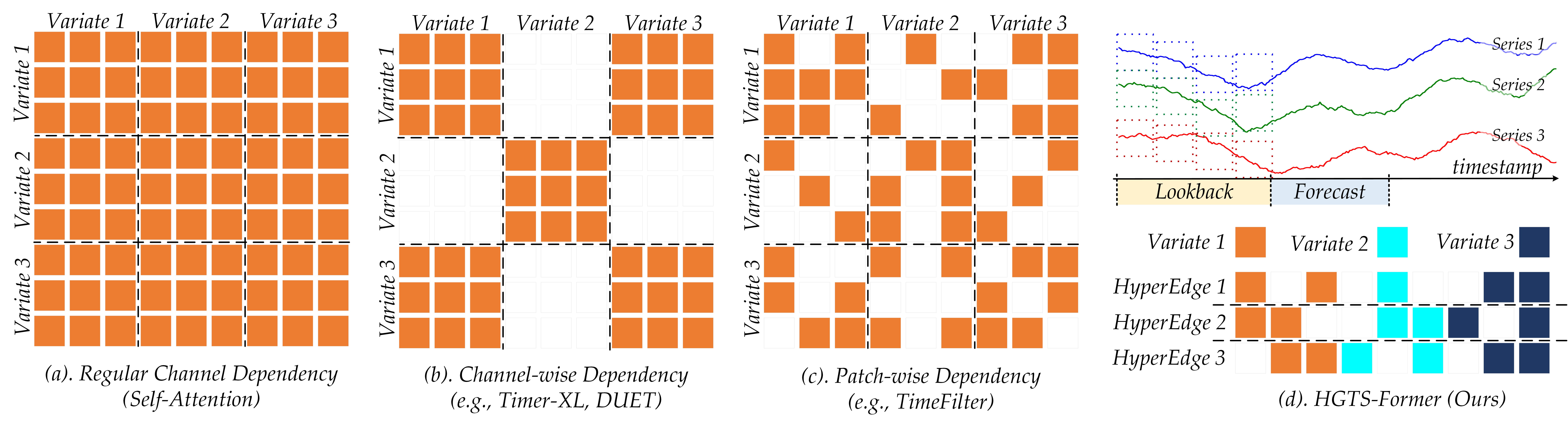}
\caption{Comparison between existing time series models and our newly proposed Hierarchical Hypergraph Transformer.} 
\label{fig:firstIMG}
\end{figure*}

To address the aforementioned issues, existing approaches model inter-variable correlations to solve multivariate time series problems. For example, Liu et al. adopt self-attention-based Transformer~\cite{liu2023iTransformer, liu2024timer-xl, zhang2025timesbert} to model the time series data, as shown in Fig.~\ref{fig:firstIMG} (a). iTransformer~\cite{liu2023iTransformer} treats each independent time series as a token and captures multivariate correlations through an attention mechanism. Timer-XL~\cite{liu2024timer-xl} utilizes a decoder-only architecture, and TimesBERT~\cite{zhang2025timesbert} adopts a BERT-style foundation model for multivariate time series forecasting. These methods are built upon a strong assumption that all channels are correlated with each other. Nevertheless, in practical scenarios, many channels do not exhibit significant correlations, as shown in Fig.~\ref{fig:firstIMG} (b). More in detail, DUET~\cite{qiu2024duet} learns the correlation between different channels in the frequency domain through channel clustering. CCM~\cite{chen2024similarity} uses a prototype-based learning approach to capture correlations between channels. To further extract fine-grained temporal correlations, some studies have shifted to patch-based modeling strategies, e.g., the TimeFilter~\cite{hu2025timefilter} uses Graph Neural Networks (GNN) to capture the spatiotemporal correlations between time-series tokens, as shown in Fig.~\ref{fig:firstIMG} (c). Some hypergraph-based methods have also shown excellent performance, e.g., MSHyper~\cite{shang2024mshyper} utilizes a multi-scale fixed hypergraph to simulate high-order pattern interactions, and Ada-MSHyper~\cite{shang2024ada} further uses a learnable hypergraph to model implicit inter-group interactions.

Despite significant improvements having been achieved, their performance is still limited by: 
(1). Traditional GNN models based on ordinary graph structures can only model binary relationships (each edge connects exactly two nodes), making it difficult to capture high-order interactions. Their message-passing mechanisms are limited to local pairwise aggregation, which restricts their expressive power and generalization capability in complex relational scenarios. 
(2). Hypergraph neural networks for MTS data suffer from limited receptive fields due to stacking layers to capture long-range dependencies, making it difficult to model the importance differences among nodes and hyperedges, and resulting in a weak capability for capturing dynamic or heterogeneous relationships.
Thus, it is natural to raise the following question: 
\textit{is it possible to design a hypergraph-based architecture with global receptive fields and adaptive attention mechanisms to overcome the limitations of both standard GNNs and existing hypergraph models for multivariate time series analysis?}

In this paper, we propose a novel hierarchical hypergraph time series Transformer network to address the aforementioned question, termed \textbf{HGTS-Former}. The key insight of this paper lies in our introduction of hypergraphs to model high-order dependencies in multivariate data, along with a general hierarchical attention architecture that captures both fine-grained and coarse-grained global information within and among variables, as shown in Fig.~\ref{fig:firstIMG} (d). Given the multivariate time series data, we first standardize its distribution and map it into the shared feature space. Then, we adopt Multi-Head Self-Attention (MHSA) to enhance the temporal representation of each channel. Hierarchical hypergraph is designed to aggregate the potential temporal patterns within variables and fine-grained soft channel correlations between variables. Then, the hyperedge information is converted into node features via the Edge2Node module and fed into the feed-forward networks to output the final representation. An overview of our proposed HGTS-Former is illustrated in Fig.~\ref{fig:framework}.

To thoroughly and comprehensively validate the effectiveness of our model, in addition to conventional public datasets, we also propose the \textbf{EAST-ELM640}, a novel benchmark dataset tailored for the multivariate analysis of nuclear fusion, specifically for the classification of Edge Localized Modes (ELMs). This dataset is compiled from 640 plasma discharges sourced from experimental campaigns of the Experimental Advanced Superconducting Tokamak (EAST) facility. To guarantee high data quality, all discharges undergo manual review by domain experts in the fusion field. Each discharge in the dataset contains 18 diagnostic signals that characterize plasma behavior, and the dataset is rationally partitioned into three subsets for model development and evaluation: 448 shots for training, 96 shots for validation, and 96 shots for testing. We retrain and benchmark nine representative recent time-series models on the EAST-ELM640 dataset, establishing a reliable baseline and laying a solid foundation for future research in this emerging field.

To sum up, the main contributions of this paper can be summarized as follows: 

1). We propose a novel hierarchical hypergraph Transformer, termed HGTS-Former, to simultaneously capture global long-range dependencies and high-order relational patterns from multivariate time series.

2). We propose the intra- and inter-hypergraph attention aggregation layers to flexibly extract and model high-order and global dependencies of various time series.  

3). We propose a large-scale benchmark dataset for the Edge Localized Modes (ELMs) recognition for nuclear fusion, termed EAST-ELM640, which contains 448, 96, and 96 shots for training, validation, and testing, respectively. 

4). Extensive experiments on nine benchmark datasets for long-term and short-term forecasting, four datasets for anomaly detection, and the newly proposed EAST-ELM640 nuclear fusion dataset for edge localized modes recognition, fully validated the effectiveness of our proposed hypergraph Transformer networks.

\textit{The rest of this paper is organized as follows:} 
In section~\ref{sec::relatedWorks}, we introduce the most relevant related works from self-attention and Transformer, time series analysis, and hypergraph neural networks. In section~\ref{sec::method}, we dive into the details of our proposed approach, with a focus on the review of graph and hypergraph, overview, input encoding, detailed network architectures, and loss function. In section~\ref{sec::experiments}, we conduct the experiments to validate the effectiveness of our model on multiple multivariate time series tasks. In section~\ref{sec::conclusion}, we conclude this paper and propose possible research directions for future work.

\section{Related Works} \label{sec::relatedWorks}

In this section, we introduce the related works with a focus on Self-Attention and Transformer, Time Series Analysis, Hypergraph Neural Network. More related works can be found in the following surveys~\cite{wang2023MMPTMSurvey}. 

% and paper list~\footnote{\url{https://github.com/Event-AHU/Time_Series_Analysis}}.

\subsection{Self-Attention and Transformer}
The Transformer architecture~\cite{vaswani2017attention} has brought about a paradigm shift in various fields~\cite{wang2023visevent, wang2024eventVOT, wang2025sstformer, wang2025PromptPAR}.   
By leveraging the self-attention mechanism, it overcomes the limitations of traditional recurrent neural networks (RNNs) in handling long-range dependencies. Self-attention allows the model to compute the attention weights between all positions in the input sequence, enabling it to capture global information more effectively. Since its advent, numerous Transformer-based models have emerged. For example, BERT~\cite{devlin2019bert} has shown remarkable performance in natural language processing (NLP) tasks. In the computer vision domain, Vision Transformer (ViT)~\cite{dosoViTskiy2020image} extends the Transformer architecture to image classification by treating image patches as sequence elements. With the success of Transformers in NLP and computer vision, many Transformer-based time series models have emerged in the field of time series analysis, such as iTransformer~\cite{liu2023iTransformer}, PatchTST~\cite{nie2022time}. 
% In time series forecasting, Transformer-like architectures have also been explored. Models like Informer~\cite{zhou2021informer} incorporate self-attention mechanisms to capture long-term temporal dependencies in time series data, achieving better performance compared to traditional methods. 

Transformers have been extensively applied to time series forecasting. Initial Transformer-based forecasters primarily focused on long-term forecasting~\cite{li2019enhancing,lim2021temporal}, with efforts to reduce the complexity of the original attention mechanism through designs such as Fourier-enhanced structures~\cite{zhou2022fedformer}, and pyramid attention modules~\cite{liu2021pyraformer}, though most relied on point-wise attention and overlooked the potential of patches. Afterwards, the field has seen a shift towards pre-trained large models~\cite{das2024decoder, woo2024unified, ansari2024chronos}, with decoder-only Transformers gaining attention. Building on these advancements, Timer-XL~\cite{liu2024timer-xl} emerges as a causal decoder-only Transformer for unified time series forecasting, generalizing next token prediction to multivariate scenarios. Zhu et al.~\cite{zhu2023mr} proposed a Multi-Resolution Transformer (MR-Transformer) for multivariate time series prediction, which models multivariate time series from two dimensions: temporal resolution and variable resolution. Waveformer~\cite{wei2025wavelet} improved the precision of time series prediction under seasonal-trend decomposition methods by a novel wavelet attention mechanism. 
Inspired by these works, we propose a novel hypergraph Transformer architecture for time series analysis.

\subsection{Time Series Analysis}  
Multivariate time series forecasting is a complex task that aims to predict future values of multiple variables considering their interdependencies. Traditional methods such as the autoregressive integrated moving average (ARIMA) model and its extensions for multiple variables~\cite{nelson1998time,zhang2003time,wang2013arima} have been widely used. However, these linear models often struggle to capture non-linear relationships and complex temporal patterns in the data. With the development of machine learning and deep learning, deep neural network-based models like Stacked Long Short-Term Memory (S-LSTM)~\cite{bhanja2020deep} have been proposed to improve the accuracy of multivariate time series forecasting.
MSA-GCN~\cite{chen2024multistage} utilized multistage graph convolutional networks to learn the hidden correlation from heterogeneous MTS for accurate data imputation.
Channel Independent (CI)~\cite{zeng2023Transformers,dai2024periodicity,nie2022time} approaches in time series analysis treat each variable in a multivariate time series independently. Within the PatchTST~\cite{nie2022time} framework, each univariate series from a multivariate time series is fed independently into the Transformer backbone, with each generating its own attention maps and prediction results while sharing the same Transformer weights. In contrast, Channel Dependent (CD)~\cite{zhang2023crossformer,yu2024reViTalizing,huang2023crossgnn} strategies focus on capturing the dependencies between variables. CrossGNN~\cite{huang2023crossgnn} designs a Cross-Variable GNN that distinguishes between homogeneous and heterogeneous relationships among variables, selecting nodes with high correlation scores as positive neighbors and those with low scores as negative neighbors.
% However, CI approaches suffer from a key limitation: by design, they neglect potential inter-channel correlations, thereby failing to exploit valuable covariate information that could enhance forecasting performance. On the other hand, CD strategies, while aiming to model the full dependency structure, tend to incorporate irrelevant or noisy relationships among variables, which can undermine model generalization and robustness.

Recent efforts~\cite{chen2024similarity,qiu2024duet,liu2025timecheat} have sought to leverage both the advantages of CI and CD strategies. For instance,  Chen et al.~\cite{chen2024similarity} introduce a Channel Clustering Module (CCM) that dynamically groups channels with intrinsic similarities, utilizes cluster information instead of individual channel identities, and employs a cluster-aware Feed Forward mechanism to integrate the strengths of both strategies. However, their coarse-grained designs are often insufficient to capture the dynamic and evolving nature of interactions. To address this limitation, TimeFilter~\cite{hu2025timefilter} aims to customize the required dependency relationships for each segment period through a fine-grained filtering mechanism. Nonetheless, TimeFilter's reliance on standard pairwise graphs and indirect inter-variable modeling fundamentally limits its capacity to capture complex higher-order group dependencies and express nuanced channel-wise correlations.

\subsection{Hypergraph Neural Network} 
Hypergraph neural networks (HGNNs), serving as a generalized variant of GNNs, have been employed across diverse domains. Hypergraphs allow hyperedges to connect more than two vertices, enabling them to represent more complex relationships. Within the realms of machine learning and deep learning, HGNNs~\cite{huang2009video,sawhney2021stock,xu2022groupnet} have been developed to handle data with such sophisticated interconnections. Specifically, GroupNet~\cite{xu2022groupnet} constructs a trainable multiscale hypergraph to capture both pairwise and group-wise interactions across multiple group sizes, and employs a three-element representation format for end-to-end learning and explicit relational reasoning in multi-agent trajectory prediction. HyperGCN~\cite{yadati2019hypergcn} laid the foundation for applying hypergraph convolution in relational data, demonstrating that hypergraphs can better model group-wise correlations compared to traditional GNNs. This insight was extended to time series forecasting, where MSHyper~\cite{shang2024mshyper} proposed a multi-scale hypergraph Transformer to model high-order pattern interactions, using rule-based hyperedges to connect time steps with inherent periodicity. However, such methods rely on predefined hypergraph structures, which may fail to capture implicit temporal correlations and introduce noise in complex scenarios. Notably, Ada-MSHyper~\cite{shang2024ada} further advances this direction by proposing an adaptive multi-scale hypergraph Transformer that models implicit group-wise interactions across scales and differentiates temporal variations, avoiding over-reliance on fixed structures. 
However, most of these studies rely on traditional HGNN methods, which may have limitations in handling large-scale data and complex temporal patterns. In our work, we introduce a novel approach that uses a hierarchical hypergraph structure in combination with a Transformer-like architecture to better model the fine-grained dynamic channel correlations and aggregate latent temporal patterns in time series data, without resorting to traditional HGNN methods.

\section{Our Proposed Approach} \label{sec::method}

\begin{figure*}
    \centering
    \includegraphics[width=1\linewidth]{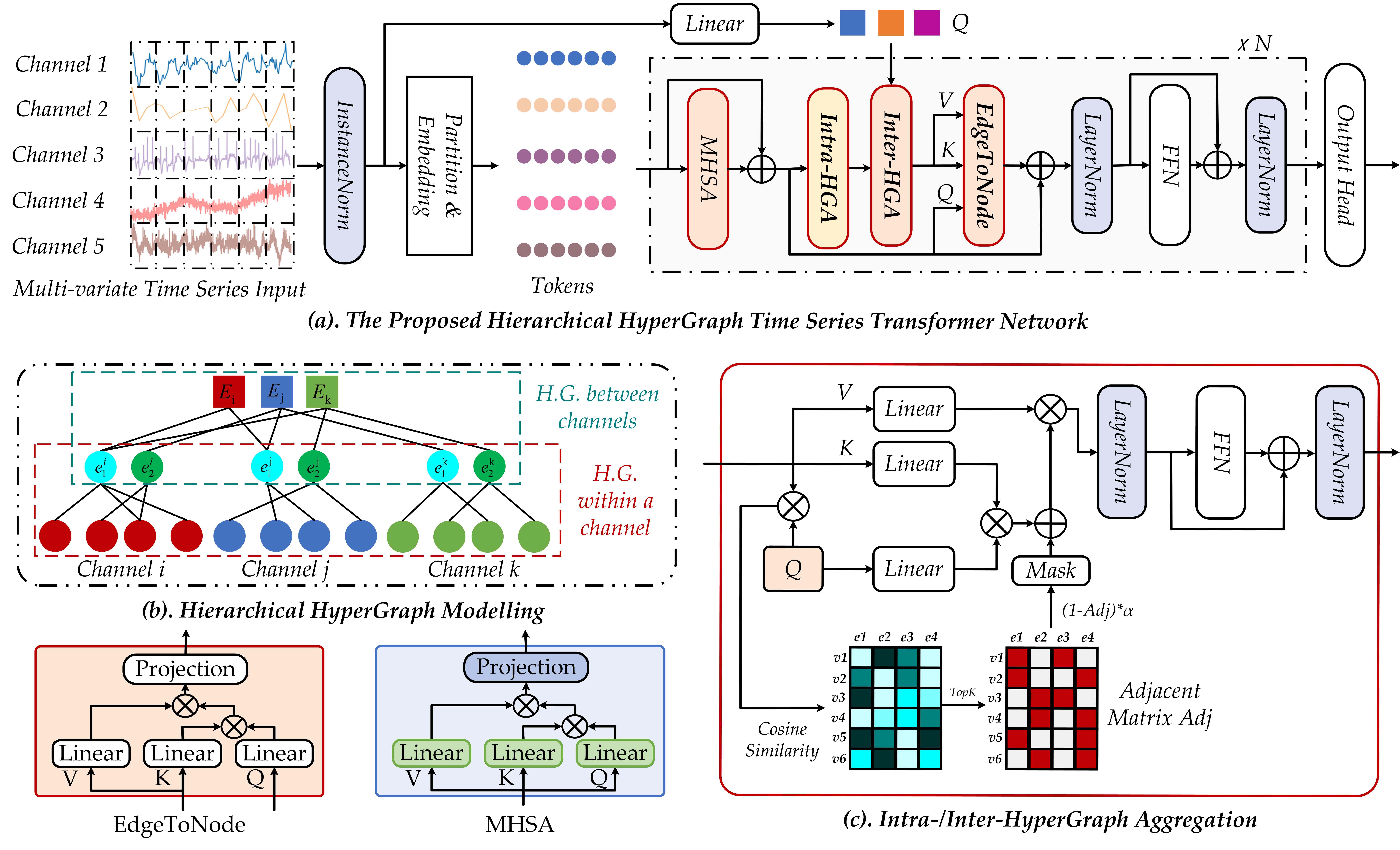}
    \caption{An overview of our proposed HyperGraph Time Series Transformer. We use the Intra-/Inter-HGA module to construct the hierarchical hypergraph in (b) and perform fine-grained aggregation within variables and coarse-grained aggregation between variables. }
    \label{fig:framework}
\end{figure*}

\subsection{Preliminaries: Graph and HyperGraph} 
Graph is one of the most basic data structures. A graph is usually defined as $\mathcal{G}=\{\mathcal{V},\mathcal{E}\}$, where $\mathcal{V}=\{v_1, v_2, ..., v_n\}$ is the vertex set, where $n$ is the number of vertices, $\mathcal{E}=\{e_1, e_2,..., e_m\}$ is the edge set, where $m$ is the number of edges, and each edge can only connect two vertices. The relationship between nodes can be represented as an adjacency matrix 
% $A \in \mathbb{R}^{N \times N }$
$\mathbf{A} \in  \mathbb{R}^{N \times N}$, where $N$ is the number of nodes. 
The adjacency matrix $\mathbf{A}$ can be expressed as below: 
\begin{equation}
\mathbf{A}_{ij} = 
\begin{cases} 
1, & v_i,v_j \in e_{ij} \\ 
0, & v_i,v_j \notin e_{ij} 
\end{cases}
\label{eq:adjacency_matrix}
\end{equation}

Unlike ordinary graphs depicting direct neighbor relationships, hypergraphs focus on modeling higher-order relationships between nodes. The fundamental reason is that each edge of a graph can only connect two nodes, while the hyperedge of a hypergraph can connect two or more nodes, thereby improving the ability to represent high-order interactions. The topological relationship of a hypergraph can be represented as an incidence matrix $\mathbf{H} \in  \mathbb{R}^{N \times M}$, where \textit{N} represents the number of nodes and \textit{M} represents the number of hyperedges. The matrix $\mathbf{H}$ is shown in the following Eq.~\ref{eq:H_matrix}: 
\begin{equation}
\mathbf{H}_{mn} = 
\begin{cases} 
1, & V_n \in e_m \\ 
0, & V_n \notin e_m 
\end{cases}
\label{eq:H_matrix}
\end{equation} 

Following the definition of the incidence matrix $\mathbf{H}$, let $\mathbf{D}_v$ and $\mathbf{D}_e$ denote the diagonal matrices representing the vertex and hyperedge degrees, respectively. 
The message passing process within a hypergraph is characterized by a "node-edge-node" transformation, in which node features are first gathered to form hyperedge features via $\mathbf{H}^\top$ and subsequently aggregated back to update node representations through $\mathbf{H}$. 
Formally, the propagation rule for the $l$-th layer is formulated as:
\begin{equation}
\mathbf{X}^{(l+1)} = \sigma \left( \mathbf{D}_v^{-1/2} \mathbf{H} \mathbf{W} \mathbf{D}_e^{-1} \mathbf{H}^\top \mathbf{D}_v^{-1/2} \mathbf{X}^{(l)} \mathbf{\Theta}^{(l)} \right)
\label{eq:hgnn_propagation}
\end{equation}
where $\mathbf{X}^{(l)}$ denotes the node feature matrix at the $l$-th layer. The diagonal matrix $\mathbf{W}$ contains hyperedge weights used to modulate the significance of different high-order correlations , while $\mathbf{\Theta}^{(l)}$ represents the learnable filter parameters that project node features into a higher-level representation space. Finally, $\sigma(\cdot)$ denotes the non-linear activation function. This mechanism enables the framework to effectively exploit complex high-order data relationships that exceed the modeling capacity of traditional pairwise graph structures.

\subsection{Overview}  
Fig.~\ref{fig:framework} illustrates the structural framework of HGTS-Former, whose core innovation lies in adopting a two-layer hypergraph structure to capture potential temporal patterns and model fine-grained dynamic channel correlations. 
Specifically, we first standardize the distribution of time series through the InstanceNorm layer, and then use the Embedding layer to divide the patches and map them to the shared feature space. Then, the MHSA module is adopted to enhance the temporal representation of each patch. We also utilize hierarchical hypergraphs to complete the aggregation of potential temporal patterns within variables and fine-grained soft channel correlations between variables. The model converts the hyperedge information into node features through the EdgeToNode module, and finally inputs the processed features into the feedforward neural network and the downstream task output head to obtain the final prediction results. 
It is worth noting that in the aggregation process of the hypergraph, we only use sparsity and Transformer~\cite{vaswani2017attention} instead of traditional HGNNs~\cite{feng2019hypergraph}.

In summary, the core logic of HGTS-Former lies in implementing a hierarchical hypergraph aggregation mechanism, which captures the complex dependencies of multivariate temporal series by cascading hypergraph operators at different spatial scales. 
This hierarchical aggregation process can be highly abstracted into the following three stages of function composition:
\begin{equation}
\mathbf{X}^{(l+1)} = \Psi \left( \Phi_{inter} \left( \Phi_{intra} \left( \mathbf{X}^{(l)}, \mathcal{H}_{intra} \right), \mathcal{H}_{inter} \right), \mathbf{X}^{(l)} \right)
\label{eq:general_framework}
\end{equation}
where $\mathbf{X}^{(l)}$ and $\mathbf{X}^{(l+1)}$ are the node features of the $i$-th layer and the $(i+1)$-th layer, respectively. $\Phi_{intra}$ represents intra-channel node aggregation, which aggregates potential temporal pattern features by constructing an intra-channel hypergraph $\mathcal{H}_{intra}$; 
$\Phi_{inter}$ represents inter-channel node aggregation, which further captures higher-order dynamic relationships across channels based on the previous stage using an inter-channel hypergraph $\mathcal{H}_{inter}$; 
it is worth noting that Transformer was used to complete $\Phi_{intra}$ and $\Phi_{inter}$ process.
$\Psi$ corresponds to the EdgeToNode module, which is responsible for projecting the finally refined higher-order hyperedge information back to the original node space to achieve accurate updates of feature representations. This hierarchical design ensures that the model can adaptively model cascading dependencies from local patterns to global couplings while maintaining the essential high-order modeling capabilities of hypergraphs.

% \subsection{Network Architecture} 

\subsection{Input Representation} 
As shown in Fig.~\ref{fig:framework}, we use the InstanceNorm layer to unify the distribution of the time series $\mathbf{X} \in {\mathbb{R}^{B \times C \times L}}$, and then use the Embedding layer to divide the 
$\mathbf{X}$ of length L into non-overlapping patches $\mathbf{X}_p \in \mathbb{R}^{B \times C \times N \times P}$, 
where $B$ is the batchsize, $P$ denotes the length of each patch, $C$ is the number of variables, 
and $N=\lceil \tfrac{L}{P} \rceil$ is the number of patches. 
Then each patch will be mapped to the shared feature space $\mathbf{X}_p \in \mathbb{R}^{B \times C \times N \times D}$, 
\begin{equation}
\mathbf{X}_{norm}=\texttt{InstanceNorm}(\mathbf{X})
\label{eq:norm}
\end{equation}
\begin{equation}
\label{eq:embedding} 
\mathbf{X}_p=\texttt{Embedding}(\mathbf{X}_{norm}) 
\end{equation}
where $D$ is the feature dimension.

To obtain a more robust feature representation, we use a Multi-Head Self-Attention (MHSA) to enhance the temporal pattern representation of each patch within each variable and to capture the complex temporal dependencies among patches within the same variable. Since the self-attention mechanism is permutation invariant, the time series is ordered in the temporal dimension. In this work, we use RoPE~\cite{su2024roformer} to inject position information into the patch token, which has been proven to be effective in Timer-XL~\cite{liu2024timer-xl}, to reflect the position relationship of each patch token in the time series. The process can be formulated as follows:
% \begin{equation}
% \mathbf{Q} = \mathbf{W}_Q\mathbf{X}_p,~~\mathbf{K} = \mathbf{W}_K\mathbf{X}_p,~~\mathbf{V} = \mathbf{W}_V\mathbf{X}_p
% \label{eq:QKV}
% \end{equation}
\begin{equation}
\hat{\mathbf{X}}_p=\texttt{MHSA}(\mathbf{X}_p)+\mathbf{X}_p
\label{eq:MHSA}
\end{equation}
where $\hat{\mathbf{X}}_p \in \mathbb{R}^{BC \times N \times D}$ denotes the enhanced node feature, $\texttt{MHSA}(\cdot)$ represents Multi-Head Self-Attention mechanism

\subsection{Hierarchical HyperGraph Time Series Blocks}

As shown in Fig.~\ref{fig:framework} (b), we leverage the hypergraph within a channel to capture latent temporal patterns within a single variable and use the hypergraph between channels to capture the dynamic dependencies between multiple variables.

\noindent $\bullet$ \textbf{Intra-HyperGraph.} 
% Fixelle et al.~\cite{fixelle2025hypergraph} proposed a hypergraph Vision Transformer that uses Transformers to map from nodes to hyperedges and from hyperedges to nodes. Inspired by it, 
In this paper, we propose the Intra-/Inter-HyperGraph Aggregation Module to complete the construction of hierarchical hypergraphs. 
As shown in Fig.~\ref{fig:framework} (c), inspired by TQNet~\cite{lin2025temporal}, we use a learnable hyperedge matrix $\mathbf{E}_{intra} \in \mathbb{R}^{E \times D}$ as the latent pattern to adaptively capture the distribution of latent patterns within the variables for the Intra-HyperGraph, where $E$ represents the number of hyperedges and $D$ represents the feature dimension. 
Then, we calculate the cosine similarity between the node and $\mathbf{E}_{intra}$, and then obtain a confidence matrix $\mathbf{M}_{conf} \in \mathbb{R}^{BC \times E \times N}$ through the sigmoid activation function. 
We sample and generate the adjacency matrix $\mathbf{Adj} \in \mathbb{R}^{BC \times E \times N}$ of the Intra-HyperGraph through the \texttt{top-k} strategy, and finally obtain the $\mathbf{Mask} \in \mathbb{R}^{BC \times E \times N}$ matrix through the $\mathbf{Adj}$ matrix. 
The process is as follows:
\begin{equation}
\mathbf{M}_{conf}=\sigma(\texttt{Similarity}(\mathbf{E}_{intra},\mathbf{\hat{X}}_P))
\label{eq:confidence matrix}
\end{equation}
\begin{equation}
\mathbf{Adj} = \texttt{top-k}(\mathbf{M}_{conf})
\label{eq:TOPK}
\end{equation}
\begin{equation}
\mathbf{Mask} = (\mathbf{1}-\mathbf{Adj}) \times \alpha
\label{eq:mask}
\end{equation}
where $\sigma(\cdot)$ operation is the sigmoid function, $\mathbf{\hat{X}}_P^{T} \in \mathbb{R}^{BC \times N \times D}$ is the transpose of $\mathbf{\hat{X}}_P \in \mathbb{R}^{BC \times D \times N}$, 
$\alpha$ is a hyperparameter that reduces the influence of non-edge nodes on the aggregation process.
% $\alpha = -1e9$.

Finally, we use the cross attention mechanism and $\mathbf{Mask}$ to complete the aggregation of nodes to hyperedges. Unlike traditional HGNNs, when performing aggregation, the attention mechanism calculates the contribution of each node to each hyperedge. This method can avoid aggregating unnecessary redundant information, and this process is completely adaptive. After the aggregation is completed, the aggregated hyperedge features are sent to the LayerNorm layer and the feedforward neural network, and finally output through the skip connection and LayerNorm layer. The aggregation process is shown in the following equations:
\begin{equation}
\mathbf{Q}=\mathbf{W}_Q\mathbf{E}_{intra},~~
\mathbf{K}=\mathbf{W}_K\mathbf{\hat{X}}_P,~~
\mathbf{V}=\mathbf{W}_V\mathbf{\hat{X}}_P 
\label{eq:QKV}
\end{equation}
\begin{equation}
\mathbf{X}_{intra}=\texttt{softmax}(\frac{\mathbf{Q}\mathbf{K}^T}{\sqrt{d_k}}+\mathbf{Mask})\mathbf{V}
\label{eq:intra_agg}
\end{equation}
\begin{equation}
\hat{\mathbf{X}}_{\mathbf{intra}}=\texttt{LN}(\texttt{FFN}(\texttt{LN}(\mathbf{X}_{intra}))+\mathbf{X}_{intra})
\label{eq:intra_agg_FFN}
\end{equation}
where $\mathbf{W}_Q$, $\mathbf{W}_K$, and $\mathbf{W}_V$ denote the parameters of the query, key, and value projection layers, respectively. $\mathbf{\hat{X}}_P \in \mathbb{R}^{BC \times N \times D}$ is the node features after MHSA enhancement, $\mathbf{Mask} \in \mathbb{R}^{BC \times E \times N}$ is the attention mechanism mask and $d_k$ is the scale factor used for scaling.

In this way, we complete the construction of Intra\-HyperGraph and aggregate node information with similar temporal patterns into hyperedge $\mathbf{e}_i^j$ as shown in Fig.~\ref{fig:framework}(b), where $i,j$ refer to the $i$-th hyperedge of the $j$-th variable.

\noindent $\bullet$ \textbf{Inter-HyperGraph.} 
After completing the construction of the Intra-HyperGraph of the variable, we obtain the hypergraph within a channel as shown in Fig.~\ref{fig:framework}(b). For the Inter-HyperGraph, we regard the hyperedges after the Intra-HyperGraph as nodes of the Inter-HyperGraph, thus, we reshape $\mathbf{\hat{X}}_{intra} \in \mathbb{R}^{BC \times E \times D}$ into $\mathbf{\tilde{X}}_{intra} \in \mathbb{R}^{B \times CE \times D}$ and reshape $\mathbf{\hat{X}}_P \in \mathbb{R}^{BC \times N \times D}$ into $\mathbf{\tilde{X}}_P \in \mathbb{R}^{B \times CN \times D}$, where $C$ is the number of channels, $E$ is the hyperedge number of Intra-HyperGraph.

Different from the learnable query of Intra-HyperGraph, we use linear layers to map the original time series $\mathbf{X} \in R^{B \times C \times L}$ to obtain hyperedges of Inter-HyperGraph $\mathbf{E}_{inter} \in \mathbb{R}^{B \times C \times D}$ with global information, which we use to guide the generation of Inter-HyperGraph and the aggregation of nodes to hyperedges. In this way, we can capture the fine-grained dynamic correlation between variables. The aggregation process is the same as Intra-HyperGraph, i.e., 
\begin{equation}
\mathbf{E}_{inter}= \texttt{Linear}(\mathbf{X})
\label{eq:E_inter}
\end{equation}

\begin{equation}
\mathbf{\hat{X}}_{inter}= \texttt{Aggregation}(\mathbf{\tilde{X}}_{intra},\mathbf{E}_{inter})
\label{eq:E_inter}
\end{equation}
% \begin{equation}
% \mathbf{Q}= \mathbf{W}_Q\mathbf{E}_{inter},
% \mathbf{K}=\mathbf{W}_K\mathbf{\tilde{X}}_{intra},
% \mathbf{V}=\mathbf{W}_V\mathbf{\tilde{X}}_{intra},
% \label{eq:QKV_inter}
% \end{equation}
% \begin{equation}
% \mathbf{X}_{inter}=softmax(\frac{\mathbf{Q}\mathbf{K}^T}{\sqrt{d_k}}+\mathbf{Mask})\mathbf{V}
% \label{eq:inter_agg}
% \end{equation}
% \begin{equation}
% \mathbf{\hat{X}}_{inter}=\texttt{LN}(\texttt{FFN}(\texttt{LN}(\mathbf{X}_{inter}))+\mathbf{X}_{inter})
% \label{eq:inter_agg_FFN}
% \end{equation}
where $\mathbf{\hat{X}}_{inter} \in \mathbb{R}^{B \times C \times D}$, the $\texttt{Aggregation}(\cdot)$ is an aggregation method similar to that of Intra-HyperGraph

% $\mathbf{W}_Q,\mathbf{W}_K,\mathbf{W}_V$  are the linear layer parameters, and $d_k$ is a scale factor. 

\noindent $\bullet$ \textbf{EdgeToNode.} 
As shown in Fig.~\ref{fig:framework}, we use the Intra-HGA module and the Inter-HGA module to complete the construction of the hierarchical hypergraph, and use the Intra-HGA module to aggregate the potential temporal patterns within the variables and use the Inter-HGA module to adaptively capture the fine-grained dynamic correlations between variables. After hierarchical aggregation, we leverage the EdgeToNode module to complete the process of hyperedge to node: 
\begin{equation}
\small 
\mathbf{Q}=\mathbf{W}_Q\mathbf{\tilde{X}}_{P},~~
\mathbf{K}=\mathbf{W}_K\mathbf{\hat{X}}_{inter},~~
\mathbf{V}=\mathbf{W}_V\mathbf{\hat{X}}_{inter},
\label{eq:QKV_edge_to_node}
\end{equation}
\begin{equation}
\small 
\mathbf{{X}_{node}}=\mathbf{W}_O\texttt{softmax}(\frac{\mathbf{Q}\mathbf{K^T}}{\sqrt{d_k}})\mathbf{V}+\mathbf{\tilde{X}}_P
\label{eq:edgetonode}
\end{equation}
where $\mathbf{X}_{node} \in R^{B \times \ CN \times D}$ and $\mathbf{W}_P$ are the projection layer parameters.
Then we feed $\mathbf{X}_{node}$ into the feedforward neural network, the process is as follows:
\begin{equation}
\mathbf{\hat{X}}_{node}=\texttt{LN}(\texttt{FFN}(\texttt{LN}(\mathbf{X}_{node})))+\mathbf{X}_{node}
\label{eq:edgetonode_FFN}
\end{equation}
where $\mathbf{\hat{X}}_{node} \in R^{B \times CN \times D}$.

\subsection{Loss Function} 
For different downstream tasks, such as forecasting and imputation, we can define different output heads. 
For forecasting and imputation tasks, we use the linear layer as the output head, and the loss function is the Mean Squared Error (MSE) loss function. 
Specifically, the MSE loss function can be written as follows:
\begin{equation}
    \mathcal{L}_{MSE} =\frac{1}{T}\sum_{i=1}^{T}(\mathbf{Y}-\mathbf{\hat{Y}})^2
\label{eq:loss}
\end{equation}
where $T$ is the forecasting or imputation length, $\mathbf{Y} \in R^{B \times C \times L}$ is the ground truth. 
More in detail, 
\begin{equation}
\mathbf{\hat{Y}}=\texttt{RevIN}(\texttt{Reshape}(\mathbf{W}_O\mathbf{\hat{X}}_{node}))
\label{eq:output}
\end{equation}
where $\mathbf{\hat{Y}} \in R^{B \times C \times L}$ is the final output, $\mathbf{W}_O$ are linear parameters and $\texttt{RevIN}(\cdot)$ operation is reverse normalization.

\begin{table*}
\centering
\caption{Full multivariate forecasting results: we conduct a rolling forecast with a single model trained on each dataset (lookback length is 672) and accomplish four forecast lengths in \{96, 192, 336, 720\}.}
\label{tab:forecast}
\resizebox{\textwidth}{!}{%
\begin{tabular}{cc|cc|cc|cc|cc|cc|cc|cc|cc|cc}
\toprule
\multicolumn{2}{c|}{} 
& \multicolumn{2}{c|}{\textbf{HGTS-Former}} 
& \multicolumn{2}{c|}{\textbf{Timer-XL}} 
& \multicolumn{2}{c|}{\textbf{Timer}} 
& \multicolumn{2}{c|}{\textbf{UniTST}} 
& \multicolumn{2}{c|}{\textbf{iTransformer}} 
& \multicolumn{2}{c|}{\textbf{DLinear}} 
& \multicolumn{2}{c|}{\textbf{PatchTST}} 
& \multicolumn{2}{c|}{\textbf{TimesNet}} 
& \multicolumn{2}{c}{\textbf{Stationary}} 
\\
\multicolumn{2}{c|}{} 
& \multicolumn{2}{c|}{\underline{(Ours)}} 
& \multicolumn{2}{c|}{\underline{(2025)}} 
& \multicolumn{2}{c|}{\underline{(2024c)}} 
& \multicolumn{2}{c|}{\underline{(2024a)}} 
& \multicolumn{2}{c|}{\underline{(2023)}} 
& \multicolumn{2}{c|}{\underline{(2023)}}
& \multicolumn{2}{c|}{\underline{(2022)}} 
& \multicolumn{2}{c|}{\underline{(2022)}} 
& \multicolumn{2}{c}{\underline{(2022b)}} 
\\
\multicolumn{2}{c|}{\multirow{-3}{*}{\textbf{Datasets}}}   
&MSE & \multicolumn{1}{c|}{MAE}   
&MSE & \multicolumn{1}{c|}{MAE} 
&MSE & \multicolumn{1}{c|}{MAE}   
&MSE & \multicolumn{1}{c|}{MAE} 
&MSE & \multicolumn{1}{c|}{MAE}   
&MSE & \multicolumn{1}{c|}{MAE} 
&MSE & \multicolumn{1}{c|}{MAE}   
&MSE & \multicolumn{1}{c|}{MAE} 
&MSE & \multicolumn{1}{c}{MAE}
\\ \midrule
\multicolumn{1}{c|}{}                               & \multicolumn{1}{c|}{96}     &0.375  &\textbf{0.393}    &\textbf{0.364}   &\underline{0.397} &0.371 &0.404  &0.379 &0.415 &0.387 &0.418 &\underline{0.369} &0.400 &0.373  &0.403 &0.452  &0.463 &0.452 &0.478                     
\\ 
\multicolumn{1}{c|}{}                               & \multicolumn{1}{c|}{192} &\textbf{0.405} &\textbf{0.414} &\textbf{0.405}                   &0.424 &\underline{0.407} &0.429 &0.415  &0.438  &0.416 &0.437               &\textbf{0.405}  &\underline{0.422} &\textbf{0.405}                        &0.425 &0.474 &0.477  &0.484 &0.510  
\\ 
\multicolumn{1}{c|}{}   & \multicolumn{1}{c|}{336} 
&\textbf{0.417} &\textbf{0.424} 
&0.427          &\underline{0.439}      
&0.434          &0.445  
&0.440          &0.454                        
&0.434          &0.450                        
&0.435          &0.445                              &\underline{0.423}    &0.440               
&0.493          &0.489                        
&0.511          &0.522                 
\\
\multicolumn{1}{c|}{}   & \multicolumn{1}{c|}{720} 
&\textbf{0.434} &\textbf{0.443}
&\underline{0.439}   &\underline{0.459}
&0.461       &0.466
&0.482       &0.482 
&0.447       &0.473   
&0.493       &0.508 
&0.445       &0.471 
&0.560       &0.534    
&0.571       &0.543           
\\
\cmidrule{2-20} 
\multicolumn{1}{c|}{\multirow{-5}{*}{ETTh1}} & \multicolumn{1}{c|}{Avg} &\textbf{0.408} &\textbf{0.419} 
&\underline{0.409} &\underline{0.430}                  
&0.418       &0.436
&0.429       &0.447
&0.421       &0.445                     
&0.426       &0.444                
&0.412       &0.435          
&0.495       &0.491               
&0.505       &0.513 
\\ \midrule
\multicolumn{1}{c|}{}  & \multicolumn{1}{c|}{96} 
&0.289 &\textbf{0.343} 
&\textbf{0.277}       &\textbf{0.343} 
&\underline{0.285} &\underline{0.344}  
&0.343       &0.398      
&0.304       &0.362               
&0.305       &0.371      
&0.289       &0.347   
&0.340       &0.374    
&0.348       &0.403      
\\
\multicolumn{1}{c|}{}  & \multicolumn{1}{c|}{192} 
&\textbf{0.346} &\textbf{0.381} 
&\underline{0.348} &\underline{0.391} 
&0.365       &0.400              
&0.376       &0.420     
&0.372       &0.407                        
&0.412       &0.439               
&0.360       &0.393           
&0.402       &0.414             
&0.408       &0.448    
\\
\multicolumn{1}{c|}{}  & \multicolumn{1}{c|}{336} 
&\textbf{0.367} &\textbf{0.399} 
&\underline{0.375}   &\underline{0.418}
&0.412       &0.440                
&0.399       &0.435 
&0.418       &0.440          
&0.527       &0.508          
&0.389       &0.420                 
&0.452       &0.452              
&0.424       &0.457                         
\\
\multicolumn{1}{c|}{} & \multicolumn{1}{c|}{720} 
&\textbf{0.385} &\textbf{0.421}
&0.409       &0.458          
&0.468       &0.487              
&0.419       &0.457 
&0.463       &0.476        
&0.830       &0.653                      
&\underline{0.398} &\underline{0.440} 
&0.462       &0.468   
&0.448       &0.476                           
\\ \cmidrule{2-20} 
\multicolumn{1}{c|}{\multirow{-5}{*}{ETTh2}}        & \multicolumn{1}{c|}{Avg} 
&\textbf{0.347} &\textbf{0.386} 
&\underline{0.352}   &0.402
&0.382       &0.418        
&0.384       &0.428     
&0.389      &0.421                   
&0.518      &0.493      
&0.359      &\underline{0.400}                      
&0.414      &0.427                       
&0.407      &0.446        
\\ \midrule
\multicolumn{1}{c|}{}   & \multicolumn{1}{c|}{96}  
&0.291&\textbf{0.336} 
&0.290  &0.341         
&\textbf{0.281}  &\underline{0.338}   
&0.289       &0.348     
&0.311       &0.365           
&0.307       &0.350     
&\underline{0.285}  &0.346              
&0.338       &0.375       
&0.414       &0.414                      
\\
\multicolumn{1}{c|}{} & \multicolumn{1}{c|}{192}
&0.335&\textbf{0.362}
&0.337       &0.369 
&\underline{0.330} &\underline{0.368}    
&0.332       &0.375     
&0.353       &0.390      
&0.337       &\underline{0.368}   
&\textbf{0.329}   &0.372                  
&0.371       &0.387 
&0.524       &0.482        
\\
\multicolumn{1}{c|}{}  & \multicolumn{1}{c|}{336} 
&\underline{0.365}&\textbf{0.382} 
&0.374        &0.392       
&0.367        &0.393 
&\underline{0.365} &0.397    
&0.387        &0.411                 
&0.366        &\underline{0.387}     
&\textbf{0.363}  &0.394        
&0.410        &0.411   
&0.541        &0.497                  
\\
\multicolumn{1}{c|}{}   & \multicolumn{1}{c|}{720} 
&\textbf{0.419} &\textbf{0.414} 
&0.437                  &0.428       
&0.432                  &0.433                
&\underline{0.421}      &0.431       
&0.452                  &0.445     
&\textbf {0.419}        &\underline{0.419}     
&\underline{0.421}      &0.426          
&0.478                  &0.450                     
&0.578                  &0.509            
\\ \cmidrule{2-20} 
\multicolumn{1}{c|}{\multirow{-5}{*}{ETTm1}}   & \multicolumn{1}{c|}{Avg} &0.353   &\textbf{0.374} 
&0.359   &0.382       
&\underline{0.352}  &0.383           
&\underline{0.352}  &0.388     
&0.376              &0.403                
&0.357              &\underline{0.381}    
&\textbf{0.349}     &0.385                  
&0.399              &0.406        
&0.514              &0.475      
\\ \midrule
\multicolumn{1}{c|}{}  & \multicolumn{1}{c|}{96}  
&0.173 &\textbf{0.254} 
&0.175        &\underline{0.257}
&0.175        &\underline{0.257}       
&\underline{0.171} &0.260  
&0.183        &0.272           
&\textbf{0.167}        &0.263  
&0.172        &0.259             
&0.187        &0.267               
&0.237        &0.306         
\\
\multicolumn{1}{c|}{}  & \multicolumn{1}{c|}{192} 
&0.234                 &\underline{0.295}
&0.242                 &0.301               
&0.239                 &0.301                                             
&\textbf{0.228}        &\textbf{0.230}      
&0.250                 &0.315 
&\underline{0.230}     &0.311         
&0.233                 &0.299             
&0.249                 &0.309     
&0.330                 &0.387       
\\
\multicolumn{1}{c|}{}  & \multicolumn{1}{c|}{336} 
&0.289&\textbf{0.331} 
&0.293                  &0.337   
&0.293                  &0.342                                            
&\underline{0.282}      &\underline{0.336} 
&0.311                  &0.356            
&0.298                  &0.361            
&\textbf{0.280}         &\textbf{0.331}         
&0.321                  &0.351           
&0.404                  &0.424
\\
\multicolumn{1}{c|}{}  & \multicolumn{1}{c|}{720} &\underline{0.372}&\underline{0.384}
&0.376            &0.390
&0.392            &0.407     
&0.380            &0.398      
&0.417            &0.419       
&0.432            &0.446                  
&\textbf{0.357}   &\textbf{0.382}          
&0.497            &0.403            
&0.525            &0.486         
\\ \cmidrule{2-20} 
\multicolumn{1}{c|}{\multirow{-5}{*}{ETTm2}}  & \multicolumn{1}{c|}{Avg} &0.267    &\underline{0.316}
&0.271               &0.322      
&0.275               &0.327     
&\underline{0.265} &\textbf{0.306}     
&0.290               &0.340               
&0.282               &0.345          
&\textbf{0.261}      &0.318            
&0.314               &0.333                 
&0.374               &0.401          
\\ \midrule
\multicolumn{1}{c|}{}  & \multicolumn{1}{c|}{96}  &\underline{0.128}&\underline{0.220}
&\textbf{0.127}                &\textbf{0.219}     
&0.129                         &0.221             
&0.130                         &0.225      
&0.133                         &0.229  
&0.138                         &0.238           
&0.132                         &0.232                                       
&0.184                         &0.288               
&0.185                         &0.287         
\\
\multicolumn{1}{c|}{}   & \multicolumn{1}{c|}{192} 
&\underline{0.147}   &\underline{0.239}
{}&\textbf{0.145}                   &\textbf{0.236}  
&0.148 &\underline{0.239}     
&0.150                              &0.244    
&0.158                              &0.258 
&0.152                              &0.251    
&0.151                              &0.250                                   
&0.192                              &0.295      
&0.282                              &0.368           
\\
\multicolumn{1}{c|}{} & \multicolumn{1}{c|}{336} 
&\underline{0.163}  &\underline{0.256}
&\textbf{0.159}                   &\textbf{0.252}   
&0.164 &\underline{0.256}         
&0.166  &0.262                           
&0.168  &0.262                            
&0.167  &0.268     
&0.171  &0.272                          
&0.200  &0.303                       
&0.289  &0.377           
\\
\multicolumn{1}{c|}{} & \multicolumn{1}{c|}{720} 
&\underline{0.200}  &\underline{0.287}
&\textbf{0.187}              &\textbf{0.277}     
&0.201                       &0.289
&0.206       &0.297      
&0.205                              &0.294   
&0.203                              &0.302    
&0.222                              &0.318                                  
&0.228                        &0.325     
&0.305                     &0.399      
\\ \cmidrule{2-20} 
\multicolumn{1}{c|}{\multirow{-5}{*}{ECL}}          & \multicolumn{1}{c|}{Avg} 
&\underline{0.159}&\underline{0.250}
&\textbf{0.155}                   &\textbf{0.246}
&0.161                            &0.251                 
&0.163                            &0.257      
&0.164                            &0.258                  
&0.165                            &0.265                 
&0.169                            &0.268   
&0.201                            &0.303 
&0.265                            &0.358          
\\ \midrule
\multicolumn{1}{c|}{}                               & \multicolumn{1}{c|}{96}  &\underline{0.347}                    &0.244
&\textbf{0.340}                       &\textbf{0.238}              
&0.348                                &\underline{0.240}                
&0.359                                &0.250         
&0.353                                &0.259          
&0.399                                &0.285          
&0.359                               &0.255   
&0.593                               &0.315         
&0.610                        &0.322      
\\
\multicolumn{1}{c|}{}                               & \multicolumn{1}{c|}{192} &\underline{0.367}                &0.253
&\textbf{0.360}                   &\textbf{0.247} 
&0.369                            &\underline{0.250}               
&0.373                            &0.257                           
&0.373                            &0.267 
&0.409                            &0.290                           
&0.377                            &0.265                                              
&0.596                            &0.317
&0.626                            &0.346                
\\
\multicolumn{1}{c|}{}   & \multicolumn{1}{c|}{336} 
&\underline{0.382}               &0.261
&\textbf{0.377}                  &\textbf{0.256}           
&0.388       &\underline{0.260}               
&0.386 &0.265      
&0.386 &0.275                      
&0.422 &0.297     
&0.393 &0.276                             
&0.600 &0.319               
&0.633 &0.352     
\\
\multicolumn{1}{c|}{} & \multicolumn{1}{c|}{720}
&\underline{0.419}   &\underline{0.282}
&\textbf{0.418}                   &\textbf{0.279}    
&0.431       &0.285             
&0.421 &0.286                          
&0.425                              &0.296  
&0.461                              &0.319  
&0.436                                               &0.305 
&0.619                        &0.335                      
&0.651                        &0.366            
\\ \cmidrule{2-20} 
\multicolumn{1}{c|}{\multirow{-5}{*}{Traffic}}      & \multicolumn{1}{c|}{Avg} &\underline{0.379}                &0.260
&\textbf{0.374}                   &\textbf{0.255}  
&0.384                            &\underline{0.259}  
&0.385                            &0.265         
&\underline{0.384}                &0.274 
&0.423                            &0.298                 
&0.391                            &0.275                                              
&0.602                            &0.322  
&0.630                            &0.347      
\\ \midrule
\multicolumn{1}{c|}{}     & \multicolumn{1}{c|}{96}  
&0.155                  &\textbf{0.200} 
&0.157                  &0.205        
&\underline{0.151}      &\underline{0.202}            
&0.152                  &0.206   
&0.174                  &0.225  
&0.169                  &0.229        
&\textbf{0.149}         &\underline{0.202}     
&0.169                  &0.228                 
&0.185                  &0.241         
\\
\multicolumn{1}{c|}{}   & \multicolumn{1}{c|}{192} 
&0.200                     &\textbf{0.242}
&0.206                     &0.250 
&\underline{0.196}         &\underline{0.245}      
&0.198                     &0.249 
&0.227                     &0.268                  
&0.211                     &0.268                          
&\textbf{0.194}            &\underline{0.245}     
&0.222                     &0.269                  
&0.286                     &0.325 
\\
\multicolumn{1}{c|}{}  & \multicolumn{1}{c|}{336} 
&\underline{0.249}            &\textbf{0.279}
&0.259                  &0.291          
&\underline{0.249} &0.288               
&0.251                              &0.291 
&0.290                              &0.309     
&0.258                              &0.306      
&\textbf{0.244}                     &\underline{0.285}   
&0.290                        &0.310                 
&0.323                        &0.347    
\\
\multicolumn{1}{c|}{}      & \multicolumn{1}{c|}{720} 
&\textbf{0.316}               &\textbf{0.325}
&0.337                        &0.344            
&0.330                        &0.344                    
&0.322                        &0.340 
&0.374                        &0.360   
&0.320                        &0.362                             
&\underline{0.317}            &\underline{0.338}      
&0.376                        &0.364                 
&0.436                        &0.401        
\\ \cmidrule{2-20} 
\multicolumn{1}{c|}{\multirow{-5}{*}{Weather}}      & \multicolumn{1}{c|}{Avg} 
&\underline{0.230}            &\textbf{0.262}
&0.240                        &0.273           
&0.232                        &0.270               
&0.231                        &0.272               
&0.266                        &0.291  
&0.239                        &0.291       
&\textbf{0.226}               &\underline{ 0.268} 
&0.264                        &0.293                   
&0.308                        &0.329            
\\ \midrule
\multicolumn{1}{c|}{}     & \multicolumn{1}{c|}{96}  
&\textbf{0.158}           &\textbf{0.204}
&\underline{0.162}        &\underline{0.221}           
&0.212                    &0.230               
&0.190                    &0.240                            
&0.183                    &0.265 
&0.193                    &0.258      
&0.168                    &0.237                                             
&0.180                    &0.272                      
&0.199                    &0.290        
\\
\multicolumn{1}{c|}{}        & \multicolumn{1}{c|}{192} 
&\textbf{0.178} &\textbf{0.219} 
&\underline{0.187}                  &\underline{0.239}     
&0.232       &0.246               
&0.223                              &0.264       
&0.205                              &0.283    
&0.214                              &0.274    
&0.189                  &0.257                
&0.199                        &0.286              
&0.243                        &0.307      
\\
\multicolumn{1}{c|}{}  & \multicolumn{1}{c|}{336} 
&\textbf{0.197} &\textbf{0.234} 
&\underline{0.205}                 &0.255      
&0.237       &\underline{0.253}                  
&0.250                              &0.283          
&0.224                              &0.299       
&0.233                              &0.291        
&0.212                  &0.277     
&0.220                        &0.301                    
&0.264                        &0.322       
\\
\multicolumn{1}{c|}{}    & \multicolumn{1}{c|}{720} 
&\textbf{0.228} &\textbf{0.259} 
&\underline{0.238}                  &0.279 
&0.252       &\underline{0.266}     
&0.292                              &0.311     
&0.239 &0.316                    
&0.246                              &0.307     
&0.240                                               &0.305   
&0.251                        &0.321              
&0.310                        &0.339    
\\ \cmidrule{2-20} 
\multicolumn{1}{c|}{\multirow{-5}{*}{Solar-Energy}} & \multicolumn{1}{c|}{Avg}
&\textbf{0.190} &\textbf{0.229}
&\underline{0.198}                  &\underline{0.249} 
&0.233       &\underline{0.249}              
&0.241                              &0.275    
&0.213                              &0.291        
&0.222                              &0.283  
&0.202                  &0.269             
&0.213                        &0.295             
&0.254                        &0.315          
\\ \midrule
\multicolumn{2}{l}{1\textsuperscript{st} Count}                         
& \multicolumn{1}{|c}{{ \textbf {15} }}& \multicolumn{1}{c|}{{ \textbf {27} }}
& \multicolumn{1}{c}{{\underline{13}}} & \multicolumn{1}{c|}{{\underline{11}}}
& \multicolumn{1}{c}{1}              & \multicolumn{1}{c|}{{0}} 
& \multicolumn{1}{c}{1}              & \multicolumn{1}{c|}{2}  
& \multicolumn{1}{c}{0}              & \multicolumn{1}{c|}{0} 
& \multicolumn{1}{c}{3}              & \multicolumn{1}{c|}{0}    
& \multicolumn{1}{c}{{11}} & \multicolumn{1}{c|}{{2}} 
& \multicolumn{1}{c}{{0}} & \multicolumn{1}{c|}{{0}} 
& \multicolumn{1}{c}{{0}} & \multicolumn{1}{c}{{0}}              
\\ \bottomrule
\end{tabular}%
}
\end{table*}

\subsection{Difference with Existing Hypergraph-based Models}
Hi-Patch~\cite{luo2025hi} is used for Irregular Multivariate Time Series prediction (IMTS). It aggregates fine-grained information within a patch through the Intra-Patch Graph Layer and aggregates information between different patches through the Inter-Patch Graph Layer. Through these two modules, local to global aggregation is completed. Inspired by this work, we use hierarchical hypergraphs to model complex relationships within a single variable and between multiple variables. 
In recent research, various methods have attempted to introduce temporal dependencies into hypergraph modeling. 
For example, HyperMixer~\cite{tian2025hypermixer} dynamically constructs hypergraph structures adapted to different temporal phases, which incorporates a router layer and multiple experts to identify distinct time periods and generate corresponding specialized hypergraphs. This design allows for flexible capture of high-order channel interactions and time-varying patterns. Nevertheless, it relies on HGNNs-based message passing with a two-phase mechanism to encode the hypergraph.
MSHyper~\cite{shang2024mshyper} proposed a multi-scale hypergraph Transformer to model high-order pattern interactions, using rule-based hyperedges to connect time steps with inherent periodicity.
Ada-MSHyper~\cite{shang2024ada} builds upon this by proposing an adaptive hypergraph learning module to construct the hypergraph. This module generates a scale-specific association matrix by calculating the similarity between node embeddings and hyperedge embeddings.  
However, these methods are all based on traditional HGNNs and rely on a fixed, two-stage message passing paradigm. Their inherent drawbacks are: high computational complexity and the message passing process is prone to over-smoothing, making it difficult for the model to capture dynamic, fine-grained relationship changes.

Unlike the traditional HGNN-based methods mentioned above, our proposed HGTS-Former fundamentally abandons the HGNN framework. Instead, we innovatively combine a sparse attention mechanism with a Transformer component to directly achieve efficient and flexible hypergraph aggregation. This design not only overcomes the shortcomings of traditional HGNN in dynamic modeling but also emphasizes fine-grained dynamic association modeling within a hierarchical hypergraph framework, thereby enabling more accurate and efficient capture of complex and evolving dependency structures in multivariate time series.

\begin{table*}[t!]
\centering
\caption{Experimental configurations of our proposed HGTS-Former, where lradj is the learning rate adjustment method.}
\label{tab:configuration}
\resizebox{\textwidth}{!}{%
\begin{tabular}{c|c|ccccccc|cccc}
\hline 
\multirow{2}{*}{\textbf{Experiment}} & \multirow{2}{*}{\textbf{Dataset}} & \multicolumn{7}{c|}{\textbf{Configuration}} & \multicolumn{4}{c}{\textbf{Training Process}} \\ \cmidrule{3-13} 
 & & \multicolumn{1}{c|}{L} & \multicolumn{1}{c|}{d\_model}    & \multicolumn{1}{c|}{d\_ff} & \multicolumn{1}{c|}{H} & \multicolumn{1}{c|}{P}  & \multicolumn{1}{c|}{edge\_num} & \multicolumn{1}{c|}{lradj}  & \multicolumn{1}{c|}{LR}     & \multicolumn{1}{c|}{Loss} & \multicolumn{1}{c|}{Batch Size} & Epochs \\ \midrule
\multirow{8}{*}{\makecell[c]{\textbf{Multivariate} \\ \textbf{Forecasting}}} 
 & ETTh1 & \multicolumn{1}{c|}{2} & \multicolumn{1}{c|}{1024} & \multicolumn{1}{c|}{2048}  & \multicolumn{1}{c|}{8}  & \multicolumn{1}{c|}{48} & \multicolumn{1}{c|}{7}         & cosine & \multicolumn{1}{c|}{0.0001} & \multicolumn{1}{c|}{MSE}  & \multicolumn{1}{c|}{32} & 10     \\ \cmidrule{2-13} 
 & ETTh2                    & \multicolumn{1}{c|}{2} & \multicolumn{1}{c|}{1024} & \multicolumn{1}{c|}{2048}  & \multicolumn{1}{c|}{8}  & \multicolumn{1}{c|}{48} & \multicolumn{1}{c|}{7}         & cosine & \multicolumn{1}{c|}{0.0001} & \multicolumn{1}{c|}{MSE}  & \multicolumn{1}{c|}{32}           & 10     \\ \cmidrule{2-13} 
 & ETTm1                    & \multicolumn{1}{c|}{1} & \multicolumn{1}{c|}{1024} & \multicolumn{1}{c|}{2048}  & \multicolumn{1}{c|}{8}  & \multicolumn{1}{c|}{96} & \multicolumn{1}{c|}{3}         & cosine & \multicolumn{1}{c|}{0.0001} & \multicolumn{1}{c|}{MSE}  & \multicolumn{1}{c|}{32}           & 10     \\ \cmidrule{2-13} 
 & ETTm2                    & \multicolumn{1}{c|}{1} & \multicolumn{1}{c|}{512}  & \multicolumn{1}{c|}{2048}  & \multicolumn{1}{c|}{8}  & \multicolumn{1}{c|}{96} & \multicolumn{1}{c|}{3}         & cosine & \multicolumn{1}{c|}{0.0001} & \multicolumn{1}{c|}{MSE}  & \multicolumn{1}{c|}{32}           & 10     \\ \cmidrule{2-13} 
 & ECL                      & \multicolumn{1}{c|}{4} & \multicolumn{1}{c|}{512}  & \multicolumn{1}{c|}{2048}  & \multicolumn{1}{c|}{8}  & \multicolumn{1}{c|}{48} & \multicolumn{1}{c|}{3}         & cosine & \multicolumn{1}{c|}{0.0001} & \multicolumn{1}{c|}{MSE}  & \multicolumn{1}{c|}{8}           & 10     \\ \cmidrule{2-13} 
 & Traffic                  & \multicolumn{1}{c|}{4} & \multicolumn{1}{c|}{512}  & \multicolumn{1}{c|}{2048}  & \multicolumn{1}{c|}{8}  & \multicolumn{1}{c|}{96} & \multicolumn{1}{c|}{4}         & cosine & \multicolumn{1}{c|}{0.0002} & \multicolumn{1}{c|}{MSE}  & \multicolumn{1}{c|}{4}           & 10     \\ \cmidrule{2-13} 
 & Weather                  & \multicolumn{1}{c|}{3} & \multicolumn{1}{c|}{512}  & \multicolumn{1}{c|}{2048}  & \multicolumn{1}{c|}{8}  & \multicolumn{1}{c|}{96} & \multicolumn{1}{c|}{4}         & cosine & \multicolumn{1}{c|}{0.0001} & \multicolumn{1}{c|}{MSE}  & \multicolumn{1}{c|}{32}           & 10     \\ \cmidrule{2-13} 
 & Solar-Energy             & \multicolumn{1}{c|}{2} & \multicolumn{1}{c|}{1024} & \multicolumn{1}{c|}{2048}  & \multicolumn{1}{c|}{8}  & \multicolumn{1}{c|}{48} & \multicolumn{1}{c|}{8}         & cosine & \multicolumn{1}{c|}{0.0002} & \multicolumn{1}{c|}{MSE}  & \multicolumn{1}{c|}{16}           & 10     \\ \midrule 
\multirow{6}{*}{\textbf{Imputation}}               & ETTm1                    & \multicolumn{1}{c|}{2} & \multicolumn{1}{c|}{256}  & \multicolumn{1}{c|}{1024}  & \multicolumn{1}{c|}{8}  & \multicolumn{1}{c|}{16} & \multicolumn{1}{c|}{21}        & cosine & \multicolumn{1}{c|}{0.002}  & \multicolumn{1}{c|}{MSE}  & \multicolumn{1}{c|}{32}           & 20     \\ \cmidrule{2-13} 
 & ETTm2                    & \multicolumn{1}{c|}{1} & \multicolumn{1}{c|}{256}  & \multicolumn{1}{c|}{1024}  & \multicolumn{1}{c|}{8}  & \multicolumn{1}{c|}{32} & \multicolumn{1}{c|}{24}        & cosine & \multicolumn{1}{c|}{0.002}  & \multicolumn{1}{c|}{MSE}  & \multicolumn{1}{c|}{32}           & 20     \\ \cmidrule{2-13} 
 & ETTh1                    & \multicolumn{1}{c|}{2} & \multicolumn{1}{c|}{256}  & \multicolumn{1}{c|}{1024}  & \multicolumn{1}{c|}{8}  & \multicolumn{1}{c|}{16} & \multicolumn{1}{c|}{24}        & cosine & \multicolumn{1}{c|}{0.002}  & \multicolumn{1}{c|}{MSE}  & \multicolumn{1}{c|}{32}           & 20     \\ \cmidrule{2-13} 
 & ETTh2                    & \multicolumn{1}{c|}{1} & \multicolumn{1}{c|}{256}  & \multicolumn{1}{c|}{1024}  & \multicolumn{1}{c|}{8}  & \multicolumn{1}{c|}{16} & \multicolumn{1}{c|}{24}        & cosine & \multicolumn{1}{c|}{0.001}  & \multicolumn{1}{c|}{MSE}  & \multicolumn{1}{c|}{32}           & 20     \\ \cmidrule{2-13} 
 & ECL                      & \multicolumn{1}{c|}{2} & \multicolumn{1}{c|}{256}  & \multicolumn{1}{c|}{512}  & \multicolumn{1}{c|}{8}  & \multicolumn{1}{c|}{16} & \multicolumn{1}{c|}{24}        & cosine & \multicolumn{1}{c|}{0.0005} & \multicolumn{1}{c|}{MSE}  & \multicolumn{1}{c|}{32}           & 20     \\ \cmidrule{2-13} 
 & Weather                  & \multicolumn{1}{c|}{1} & \multicolumn{1}{c|}{256}  & \multicolumn{1}{c|}{1024}  & \multicolumn{1}{c|}{8}  & \multicolumn{1}{c|}{16} & \multicolumn{1}{c|}{24}        & cosine & \multicolumn{1}{c|}{0.002}  & \multicolumn{1}{c|}{MSE}  & \multicolumn{1}{c|}{32}           & 20     \\ \bottomrule
\end{tabular}%
}
\end{table*}

\begin{figure*}[t]
\centering
\includegraphics[width=\textwidth]{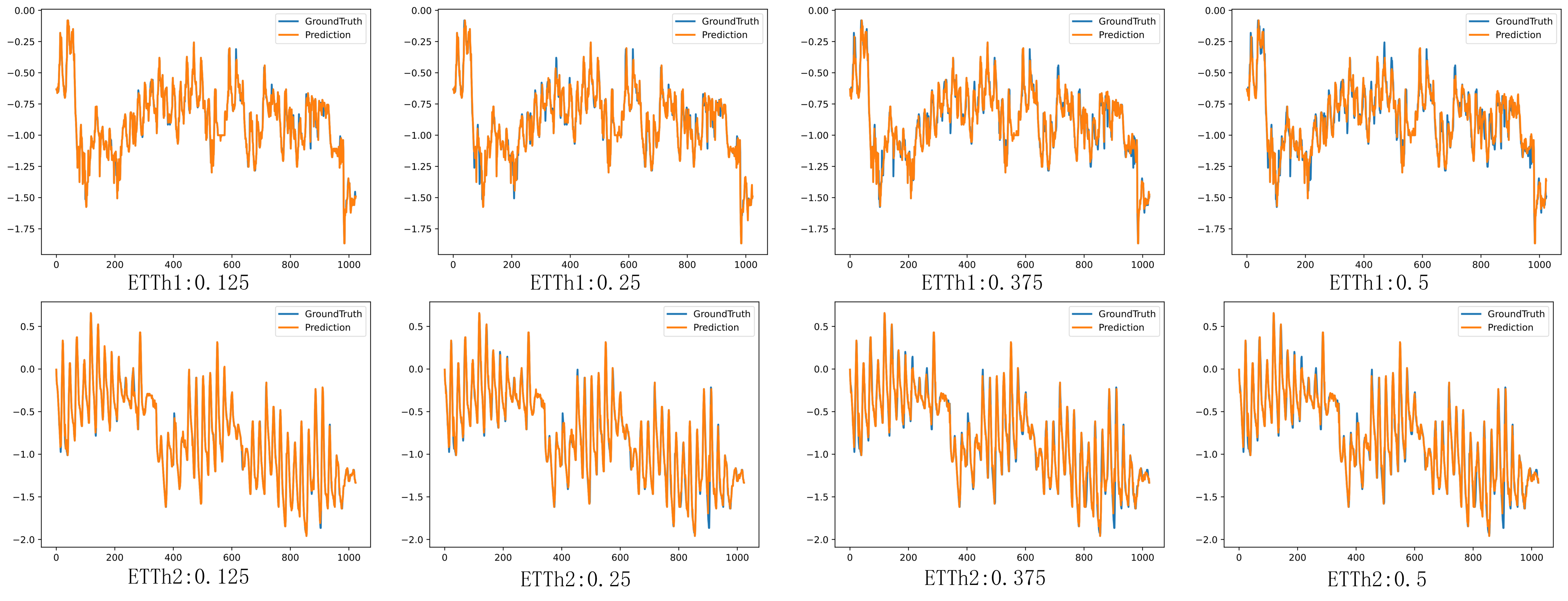}
\caption{Imputation result of our proposed model on the ETTh1 and ETTh2 datasets.} 
\label{fig:imputationResult}
\end{figure*}

\section{Experiments} \label{sec::experiments}

\subsection{Datasets and Evaluation Metric} 
For \textbf{long-term forecasting}, we conduct experiments on eight widely used public benchmark datasets: 
\textbf{ETTh1}, \textbf{ETTh2}, \textbf{ETTm1}, \textbf{ETTm2}~\cite{zeng2023Transformers}, \textbf{ECL}~\cite{wu2021autoformer}, \textbf{Traffic}~\cite{wu2021autoformer}, \textbf{Weather}~\cite{wu2021autoformer}, and \textbf{Solar-Energy}~\cite{lai2018modeling}. More in detail, 
(1) ETTh1 and ETTh2 are electricity datasets, containing 7 variables and sampled every hour. 
(2) ETTm1 and ETTm2 are electricity datasets, containing 7 variables and sampled every fifteen minutes. 
(3) ECL is an electricity dataset, containing 321 variables and sampled every hour. 
(4) Traffic is a transportation dataset, containing 862 variables and sampled every hour. 
(5) Weather is a climate dataset, containing 21 variables and sampled every ten minutes. 
(6) Solar-Energy is an energy dataset, containing 137 variables and sampled every ten minutes.

For \textbf{anomaly detection}, we evaluate our method on four widely used real-world time series anomaly detection datasets, including \textbf{MSL}~\cite{hundman2018detecting}, \textbf{SMAP}~\cite{hundman2018detecting}, \textbf{SMD}~\cite{su2019robust}, and \textbf{PSM}~\cite{abdulaal2021practical}, which originate from different application domains and contain varying numbers of variables. More in detail, 
the MSL dataset is collected by NASA from the Mars Science Laboratory rover and consists of telemetry data reflecting the operational status of onboard sensors and actuators during space missions. The original dataset is multivariate, containing 55 variables that record different physical and system-level measurements. 
The SMAP dataset is also provided by NASA and originates from spacecraft monitoring systems designed for observing soil moisture and related environmental factors. The dataset contains 25 variables in its original form, including both continuous and discrete measurements. 
The SMD dataset is collected from server machines in a large-scale Internet company and records resource utilization metrics of computer clusters over time. It includes measurements such as CPU usage, memory consumption, and other system-level indicators. The dataset is inherently multivariate and consists of 38 variables, making it a representative benchmark for multivariate time series anomaly detection in real-world industrial scenarios. 
The PSM dataset is sourced from eBay server machines and captures pooled performance metrics related to server operations under diverse workloads. It contains 25 variables describing different aspects of server behavior and exhibits a relatively high anomaly ratio, which poses additional challenges for accurate anomaly detection. The dataset is used in its full multivariate form in our experiments.

For \textbf{short-term forecasting}, we conduct experiments on the \textbf{M4 dataset}~\cite{makridakis2018m4}, which contains univariate marketing data for yearly, quarterly, monthly, and daily periods.

\begin{figure}
\centering
\includegraphics[width=1\linewidth]{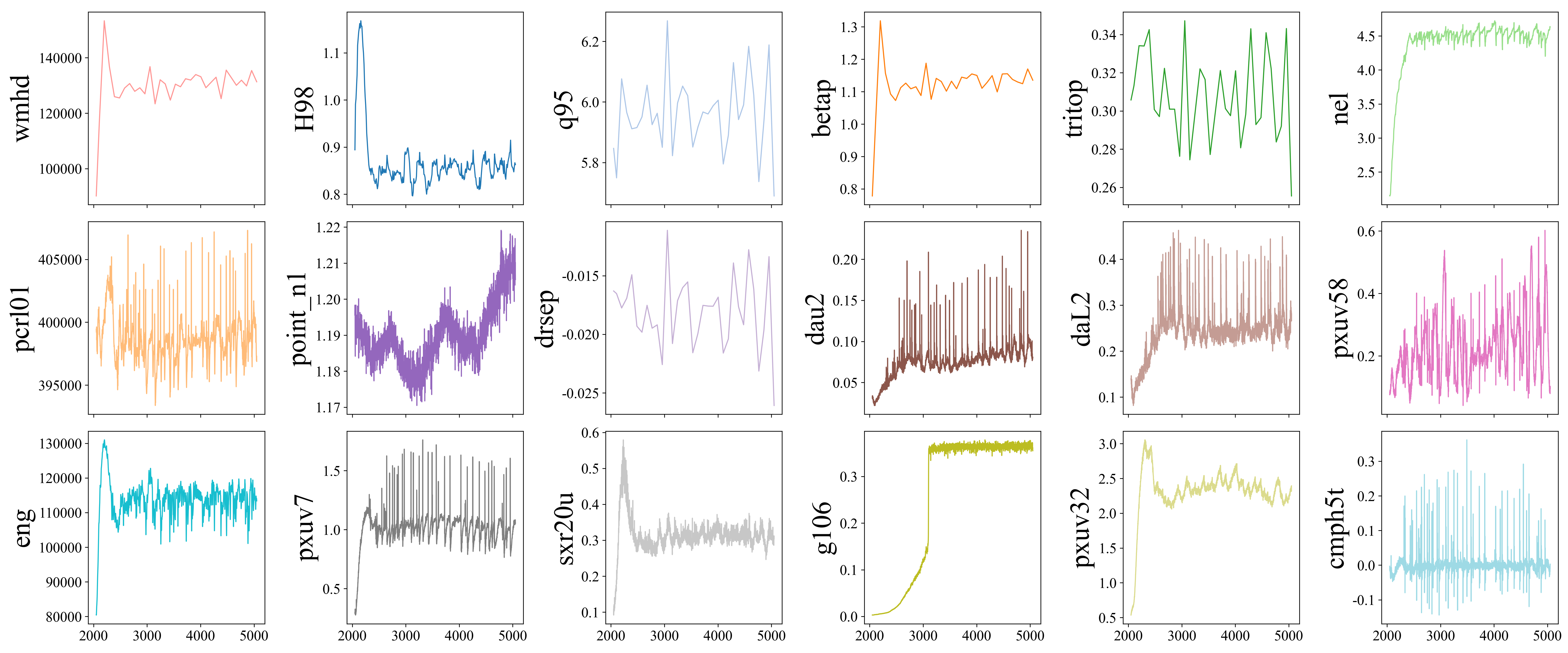}
\caption{An illustration of representative samples in the EAST-ELM640 datasets.} 
\label{fig:ELM}
\end{figure}

In addition, we also propose a new benchmark dataset for the multivariate analysis of nuclear fusion, termed \textbf{EAST-ELM640}. We compile a dataset of $640$ plasma discharges from the EAST~\footnote{\url{http://east.ipp.ac.cn/}} experimental campaigns for the classification of Edge Localized Modes (ELMs). To ensure data quality, all discharges are manually reviewed by domain experts. Each discharge comprises 18 diagnostic signals. We partition the dataset into $448$ shots for training, $96$ for validation, and $96$ for testing. Fig.~\ref{fig:ELM} illustrates the representative waveforms from a single discharge.

We choose Mean Squared Error (MSE) and Mean Absolute Error (MAE) as our evaluation metrics, where lower values represent better performance.

\subsection{Implementation Details} 
As shown in the Table~\ref{tab:configuration}, we show the core hyperparameters in the training phase, where edge num is the number of hyperedges in Intra-HyperGraph. For the construction of the hypergraph, we use \texttt{top-k} sampling, where $\texttt{top-k}=\texttt{max}(\lfloor \frac{edge~num}{3} \rfloor,1)$. In addition, Inter-graph is generated according to the same strategy. $P$ is the patch length, and $L$ is the number of the layer. We utilize Adam as our optimizer. All experiments are implemented by Pytorch on a single A800 with 80GB of memory. More details can be found in our source code.

\begin{table*}[t]
\centering
\caption{Full results of the imputation task across six datasets. To evaluate our model performance, we randomly mask \{12.5\%,25\%, 37.5\%, 50\% \} of the time points in the time series of length 1024. The final results are averaged across these 4 different masking ratios.}
\label{tab:imputation}
\resizebox{\textwidth}{!}{%
\begin{tabular}{l|cccccccccccc}
\toprule
& \multicolumn{2}{c}{ETTm1} & \multicolumn{2}{c}{ETTm2} & \multicolumn{2}{c}{ETTh1} & \multicolumn{2}{c}{ETTh2} & \multicolumn{2}{c}{ECL} & \multicolumn{2}{c}{Weather} \\
\multirow{-2}{*}{}  &MSE & \multicolumn{1}{c|}{MAE} &MSE & \multicolumn{1}{c|}{MAE} &MSE & \multicolumn{1}{c|}{MAE} &MSE & \multicolumn{1}{c|}{MAE} &MSE   & \multicolumn{1}{c|}{MAE} &MSE &MAE \\ \midrule 
\textbf{HGTS-Former} (Ours) &\textbf{0.040} & \multicolumn{1}{c|}{\textbf{0.126}} &\underline{ 0.034}    & \multicolumn{1}{c|}{\textbf{0.112}} &\textbf{0.085} & \multicolumn{1}{c|}{\textbf{0.192}} &\textbf{0.065}    & \multicolumn{1}{c|}{\underline{ 0.164}}    &\textbf{0.051}               & \multicolumn{1}{c|}{\textbf{0.146}} &\textbf{0.045} &\underline{ 0.091}    \\ \midrule
\textbf{TimeMixer++} (2025)  &\underline{0.041}    & \multicolumn{1}{c|}{\underline{ 0.127}}    &\textbf{0.024} & \multicolumn{1}{c|}{0.135}          &\underline{ 0.091}    & \multicolumn{1}{c|}{\underline{ 0.198}}    &\textbf{0.065} & \multicolumn{1}{c|}{\textbf{0.157}} &0.109                  & \multicolumn{1}{c|}{\underline{ 0.197}}    &\underline{ 0.049}    &\textbf{0.078} \\ \midrule
\textbf{TimeMixer} (2024b)   &0.072          & \multicolumn{1}{c|}{0.178}          &0.061          & \multicolumn{1}{c|}{0.166}          &0.152          & \multicolumn{1}{c|}{0.242}          &0.101          & \multicolumn{1}{c|}{0.294}          &0.142                        & \multicolumn{1}{c|}{0.261}          &0.091          &0.114          \\ \midrule
\textbf{iTransformer} (2023) &0.075          & \multicolumn{1}{c|}{0.177}          &0.055          & \multicolumn{1}{c|}{0.169}          &0.130          & \multicolumn{1}{c|}{0.213}          &0.125          & \multicolumn{1}{c|}{0.259}          &0.140                        & \multicolumn{1}{c|}{0.223}          &0.095          &0.102          \\ \midrule
\textbf{PatchTST} (2023)                            &0.097                                 & \multicolumn{1}{c|}{0.194}          &0.080 & \multicolumn{1}{c|}{0.183} &0.178 & \multicolumn{1}{c|}{0.231} &0.124 & \multicolumn{1}{c|}{0.293} &0.129 & \multicolumn{1}{c|}{0.198} &0.082 &0.149 \\ \midrule
\textbf{Crossformer} (2023) &0.081 & \multicolumn{1}{c|}{0.196} &0.111 & \multicolumn{1}{c|}{0.219} &0.201 & \multicolumn{1}{c|}{0.309} &0.206 & \multicolumn{1}{c|}{0.308} &0.125 & \multicolumn{1}{c|}{0.204} &0.150 &0.111 \\ \midrule
\textbf{FEDformer} (2022b) &0.048 & \multicolumn{1}{c|}{0.152} &0.087 & \multicolumn{1}{c|}{0.198} &0.099 & \multicolumn{1}{c|}{0.225} &0.262 & \multicolumn{1}{c|}{0.344} &0.181 & \multicolumn{1}{c|}{0.314} &0.064 &0.139 \\ \midrule
\textbf{TIDE} (2023a) &0.090 & \multicolumn{1}{c|}{0.210} &0.169 & \multicolumn{1}{c|}{0.263} &0.289 & \multicolumn{1}{c|}{0.395} &0.709 & \multicolumn{1}{c|}{0.596} &0.182 & \multicolumn{1}{c|}{0.202} &0.063 &0.131 \\ \midrule
\textbf{DLinear} (2023) &0.076 & \multicolumn{1}{c|}{0.191} &0.088 & \multicolumn{1}{c|}{0.198} &0.174 & \multicolumn{1}{c|}{0.288} &0.120 & \multicolumn{1}{c|}{0.238} &\underline{0.080} & \multicolumn{1}{c|}{0.200} &0.071 &0.107 \\ \midrule
\textbf{TimesNet} (2023) &0.049 & \multicolumn{1}{c|}{0.147} &0.035 & \multicolumn{1}{c|}{\underline{ 0.124}} &0.142 & \multicolumn{1}{c|}{0.258} &\underline{0.088} & \multicolumn{1}{c|}{0.198} &0.135 & \multicolumn{1}{c|}{0.255} &0.061 &0.098 \\ \midrule 
\textbf{MICN} (2023a) &0.059 & \multicolumn{1}{c|}{0.170} &0.109 & \multicolumn{1}{c|}{0.221} &0.127 & \multicolumn{1}{c|}{0.254} &0.179 & \multicolumn{1}{c|}{0.290} &0.138 & \multicolumn{1}{c|}{0.246} &0.075 &0.126 \\ \bottomrule
\end{tabular}%
}
\end{table*}

\begin{figure*}[t]
\centering
\includegraphics[width=1\textwidth]{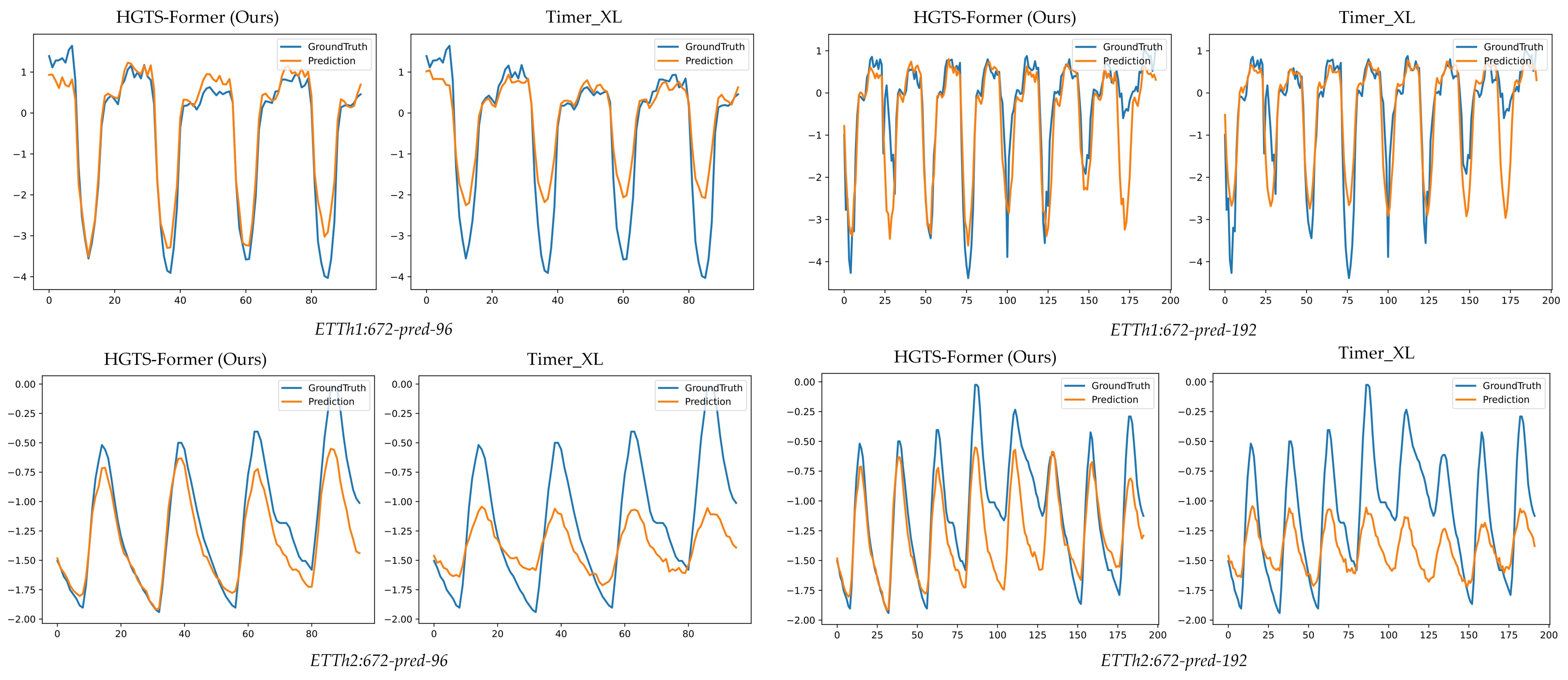}
\caption{Comparison of the predicted results between ours and Timer-XL on samples from the ETTh1 and ETTh2 datasets.} 
\label{fig:ComparTimerXL}
\end{figure*}

\begin{figure*}[t]
\centering
\includegraphics[width=1\textwidth]{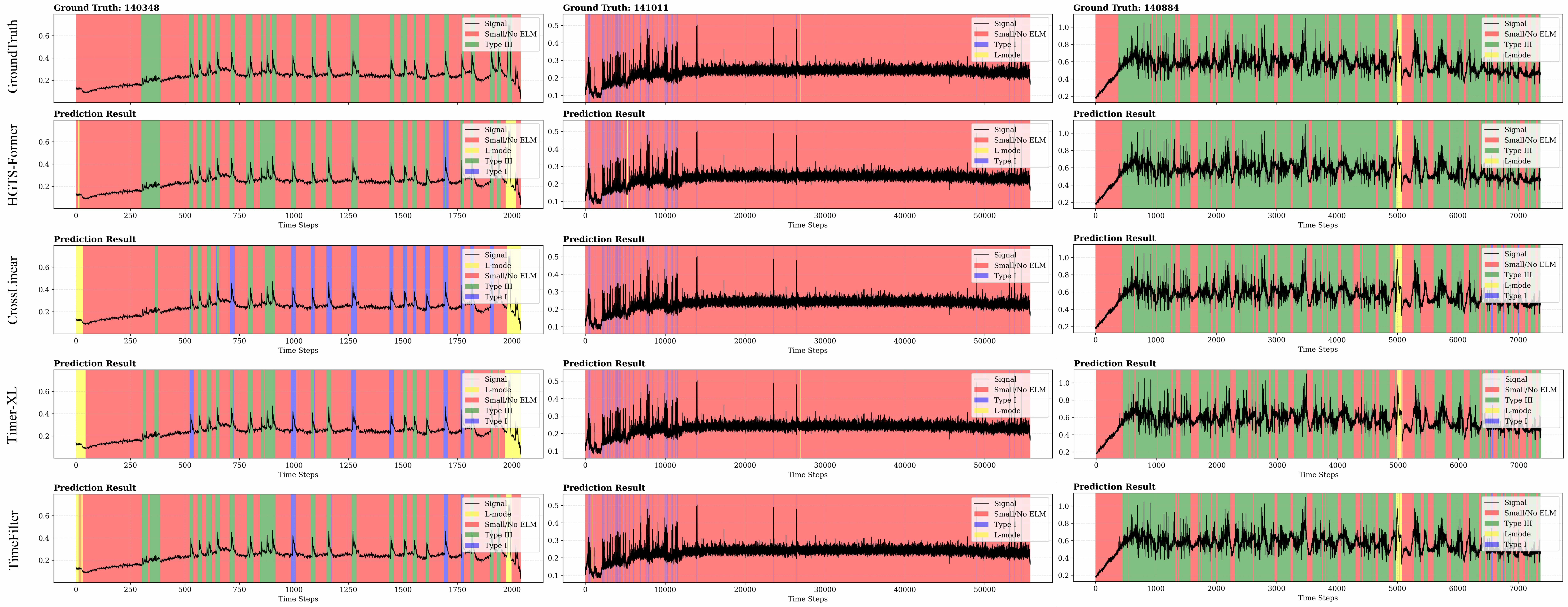}
\caption{Comparison of the ELM results between ours and others on samples from the EAST-ELM640 dataset.}  
\label{fig:ELM_visualization}
\end{figure*}

\begin{figure*}
\centering
\includegraphics[width=1\linewidth]{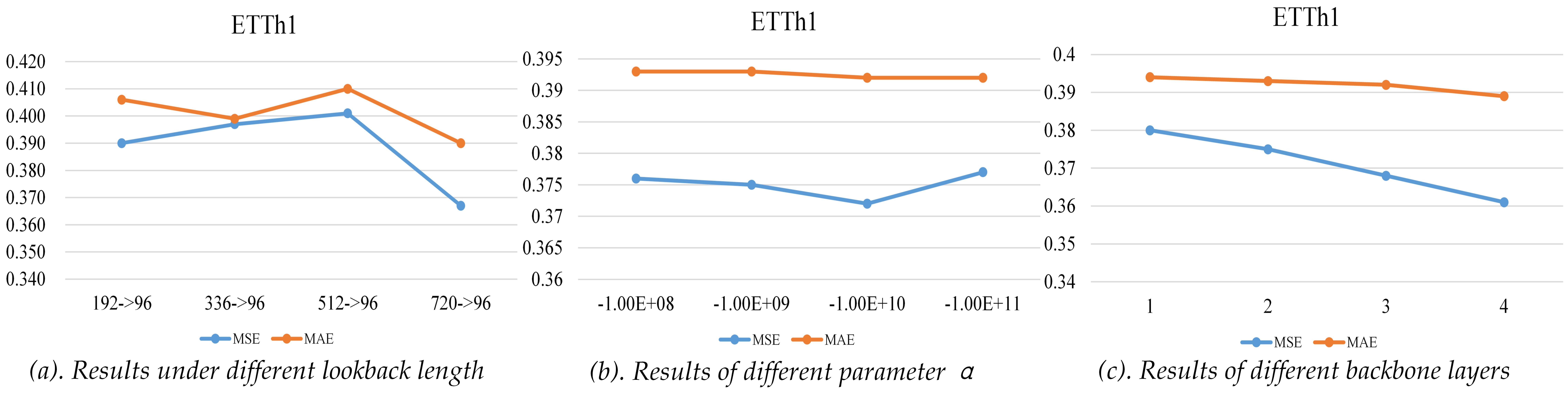}
\caption{Ablation study of different parameter configurations in our experiments.}
\label{fig:ablationStudy}
\end{figure*}

\begin{figure*}[!htp] 
\centering
\includegraphics[width=\textwidth]{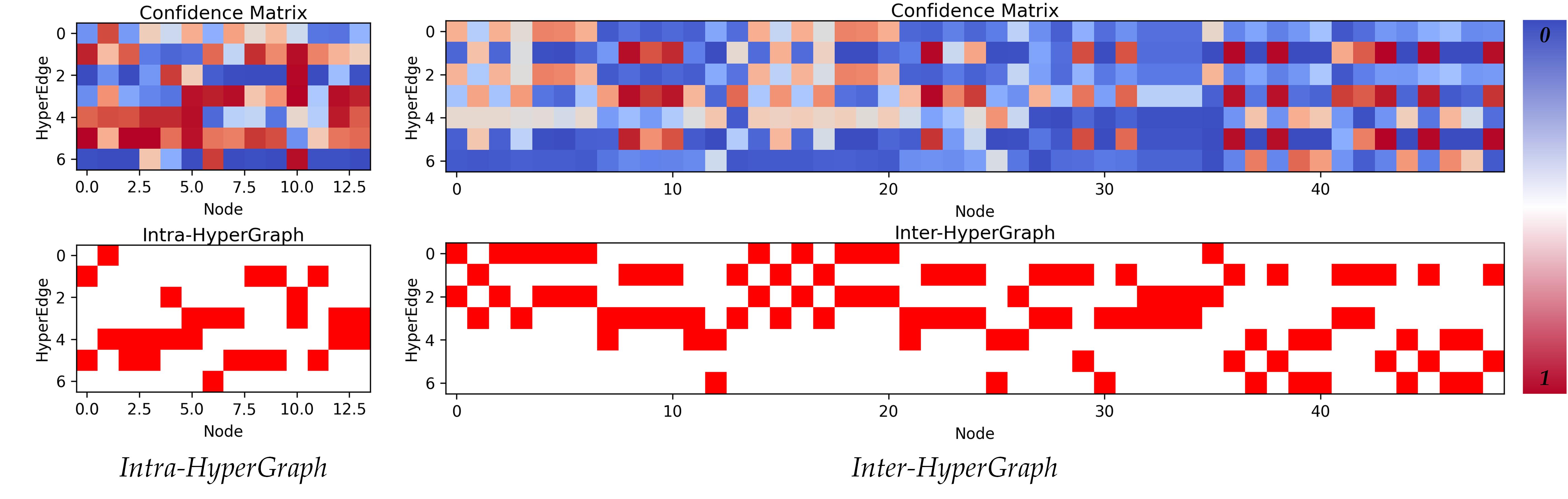}
\caption{Visualization of the Confidence Matrix and HyperGraph Visualization.}
\label{fig:hierarchicalhypergraph}
\end{figure*}

\subsection{Comparison on Public Benchmark Datasets}

\noindent $\bullet$ \textbf{Results on Long-term Forecasting Task.}
Multivariate time series forecasting is crucial for predictive performance and decision-making value in complex real-world scenarios. 
% To comprehensively evaluate the effectiveness of our model in multivariate time series forecasting, 
% we conduct experiments on eight widely used datasets, including ETTh1, ETTh2, ETTm1, ETTm2, ECL, Traffic, Weather, and Solar-Energy. 
Therefore, we train a single model on each dataset and use a rolling forecast approach (Lookback sequence length is 672) to evaluate it on four forecast steps of 96, 192, 336, and 720. 
It is worth noting that our method leverages an autoregressive method to predict the next token, thus, we only need to perform supervised training at a fine-grained level to predict arbitrary lengths adaptively.

The results in Table~\ref{tab:forecast} demonstrate the superior performance of our model in multivariate time series forecasting. 
Our model performs well in 32 forecasting tasks on 8 widely used datasets. 
It achieved 15 firsts in MSE and 27 firsts in MAE, far exceeding other advanced models, reflecting its good generalization ability and accuracy. 
As the forecast length increases, such as in tasks with forecast lengths of 336 and 720, the errors of most models increase significantly under long-term forecasts, while the performance of our model improves steadily, reflecting our model's excellent modeling ability in long-term dependencies.

\noindent $\bullet$ \textbf{Results on short-term Forecasting Task.}
We adopt the M4 dataset~\cite{makridakis2018m4}, which contains univariate marketing data for yearly, quarterly, and monthly periods, for short-term forecasting.
utilize Symmetric Mean Absolute Percentage Error (SMAPE), Mean Absolute Scaling Error (MASE), and Overall Weighted Average (OWA) to measure model performance.

As shown in the Table~\ref{tab:short-term}, our model outperforms the existing best models in all metrics.
Compared with Timer-XL~\cite{liu2024timer-xl}, the average SMAPE decreased by $0.122$, the average MASE decreased by $0.023$, and the average OWA decreased by $0.010$.
The advantages are even more pronounced compared to iTransformer~\cite{liu2023iTransformer}, Time-LLM~\cite{jin2023time}, and $S^2$IP-LLM~\cite{pan2024s}.

\begin{table*}[]
\renewcommand{\arraystretch}{1.2}
\centering
\caption{Results for the short-term forecasting task in the M4 datasets with a single variate. 
A lower SMAPE, MASE, or OWA indicates a better prediction. }
\label{tab:short-term}
\resizebox{\textwidth}{!}{ 
\begin{tabular}{cccccccccccccc}
\toprule 
\multicolumn{2}{c}{\textbf{Methods}}  
& \textbf{HGTS-Former} & \textbf{SymTime} & \textbf{Timer-XL} & \textbf{Peri-midFormer} & \textbf{S$^2$IP-LLM} & \textbf{Time-LLM} & \textbf{GPT4TS} & \textbf{TimeMixer} & \textbf{PatchTST} &\textbf{iTransformer}  & \textbf{TimesNet}    & \textbf{DLinear} \\
\multicolumn{2}{c}{\textbf{Metric}}                                                   
& \textbf{(Ours)} & (2025) & (2025)   & (2024b)   & (2024)   & (2024)   & (2023) & (2024b)   
& (2023)  & (2024b) & (2023)     & (2023)  \\ \midrule
\multirow{3}{*}{\rotatebox{90}{Yearly}}    
& \multicolumn{1}{|c|}{SMAPE}
& \textbf{13.332}  & 13.355  &13.415
& 13.483    & 14.931   
& 13.450  & 14.847  & 13.369    & 13.677   
& 13.724  & 13.463  & \underline{14.340}  \\
& \multicolumn{1}{|c|}{MASE}  
& \textbf{2.990}   & \underline{2.997}   &3.030
& 3.080     & 3.345    
& 3.184   & 3.628   & 3.009     & 3.049    
& 3.157   & 3.058   & 3.112   \\
& \multicolumn{1}{|c|}{OWA}   
& \textbf{0.784}   & \underline{0.786}   &0.792
& 0.800     & 0.878    
& 0.819   & 0.911   & 0.787     & 0.802    
& 0.817   & 0.797   & 0.830   
\\ \hline
\multirow{3}{*}{\rotatebox{90}{Quarterly}} 
& \multicolumn{1}{|c|}{SMAPE} 
& \underline{10.052}        & 10.060  & 10.131
& \textbf{10.037}    & 10.655   
& 10.671  & 10.389  & 10.131    & 10.922   
& 13.473  & 10.069  & 10.510  \\
& \multicolumn{1}{|c|}{MASE}  
& \textbf{1.170}        & 1.183   &1.182
& \textbf{1.170}     & 1.249    
& 1.276   & 1.228   & 1.186     & 1.326    
& 1.722   & \underline{1.175}   & 1.241   \\
& \multicolumn{1}{|c|}{OWA}   
& \underline{0.883}        & \textbf{0.872} & 0.891
& 0.882     & 0.939    
& 0.950   & 0.919   & 0.893     & 0.979    
& 1.240   & 0.886   & 0.930   
\\ \hline
\multirow{3}{*}{\rotatebox{90}{Monthly}}   
& \multicolumn{1}{|c|}{SMAPE} 
& \textbf{12.608}        & \textbf{12.608} & 12.782
& 12.795    & 13.012   
& 13.416  & 12.907  & 12.762    & 14.200   
& 13.674  & \underline{12.760}  & 13.382  \\
& \multicolumn{1}{|c|}{MASE}  
& \textbf{0.921}        & \underline{0.925} & 0.950
& 0.948     & 0.973    
& 1.045   & 0.954   & 0.940     & 1.111    
& 1.068   & 0.947   & 1.007   \\
& \multicolumn{1}{|c|}{OWA}   
& \textbf{0.870}        & \underline{0.872} &0.890
& 0.889     & 0.909    
& 0.957   & 0.896   & 0.884     & 1.015    
& 0.976   & 0.887   & 0.937   \\ \hline
\multirow{3}{*}{\rotatebox{90}{Others}}    
& \multicolumn{1}{|c|}{SMAPE} 
& \textbf{4.670}        & 4.941  &\underline{4.674}
& 4.912     & 5.540    
& 4.973   & 5.266   & 5.085     & 5.658    
& 5.598   & 4.995   & 5.122   \\
& \multicolumn{1}{|c|}{MASE}  
& \underline{3.217}        & 3.327 &\textbf{3.141}
& 3.260     & 8.426    
& 3.412   & 3 .595  & 3.403     & 3.626    
& 3.957   & 3.346   & 3.608   \\
& \multicolumn{1}{|c|}{OWA}   
& \underline{0.999}        & 1.045 &\textbf{0.987}
& 1.031     & 3.792    
& 1.059   & 1.121   & 1.072     & 1.167    
& 1.213   & 1.053   & 1.108   
\\ \hline
\multirow{3}{*}{\rotatebox{90}{Average}}
& \multicolumn{1}{|c|}{SMAPE} 
& \textbf{11.764}        & \underline{11.785} &11.886
& 11.897    & 12.514   
& 12.584  & 12.367  & 11.885    & 12.866   
& 13.233  & 11.888  & 12.500  \\
& \multicolumn{1}{|c|}{MASE}  
& \textbf{1.572}        & \underline{1.584}  &1.594
& 1.607     & 1.726    
& 1.763   & 1.767   & 1.598     & 1.734    
& 1.850   & 1.607   & 1.678   \\
& \multicolumn{1}{|c|}{OWA}   
& \textbf{0.845}        & \underline{0.849} &0.855
& 0.859     & 0.913    
& 0.915   & 0.918   & 0.856     & 0.928    
& 0.972   & 0.858   & 0.899   \\ \bottomrule
\end{tabular}%
}
\end{table*}

\noindent $\bullet$ \textbf{Results on Imputation Task.}  
Accurately filling missing values is crucial in time series analysis, which directly affects the accuracy of model prediction results in practical applications. 
% We use datasets from the fields of electricity and climate, and select ETTm1, ETTm2, ETTh1, ETTh2, ECL, and Weather as benchmarks. The input sequence length is 1024.

Table~\ref{tab:imputation} shows the performance of our model in filling missing values in time series on six different datasets. 
To evaluate our model, we randomly mask {12.5\%, 25\%, 37.5\%, 50\%} of the time in a time series of length 1024, and the final result is the average of these 4 different masking rates. 
The MSE and MAE of our model are significantly lower than those of other advanced models, especially on ETTm1 and ETTh1, where our model consistently outperforms other models. Compared with the second-best performing model TimeMixer++~\cite{wang2024timemixer++}, our model reduces the MSE by 15.6\% and MAE by 6.8\% on average. 
In datasets with higher complexity (such as ETTm1, ETTm2, and ECL), the error of our model is much lower than that of other models, demonstrating its ability to capture complex time series patterns. 
At random masking ratios of 12.5\% to 50\%, our model can accurately predict different degrees of missing data, demonstrating the robustness of our model.

\begin{table*}[]
\centering
\caption{Full results for the anomaly detection task. The P, R, and F1 represent the precision, recall, and F1-score (\%), respectively. F1-score is the harmonic mean of precision and recall. A higher value of P, R, and F1 indicates a better performance.} 
\label{tab:anomaly detection}
\resizebox{\textwidth}{!}{%
\begin{tabular}{ccccccccccccccc}
\toprule
\multicolumn{2}{c}{\textbf{Datasets}} & \multicolumn{3}{c}{\textbf{SMD}} & \multicolumn{3}{c}{\textbf{MSL}} & \multicolumn{3}{c}{\textbf{SMAP}} & \multicolumn{3}{c}{\textbf{PSM}} & \textbf{Avg F1} \\ \cmidrule(lr){3-5}\cmidrule(lr){6-8}\cmidrule(lr){9-11}\cmidrule(lr){12-14}
\multicolumn{2}{c}{\textbf{Metrics}}                 & P     & R     & \multicolumn{1}{c|}{F1}    & P     & R     & \multicolumn{1}{c|}{F1}    & P     & R     & \multicolumn{1}{c|}{F1}        & P     & R     & \multicolumn{1}{c|}{F1}    & (\%)   \\ \midrule
Anomaly*     & \multicolumn{1}{c|}{(2022)}  
& \underline{88.91} & 82.23 & \multicolumn{1}{c|}{\underline{85.49}} 
& 79.61 & \textbf{87.37} & \multicolumn{1}{c|}{\textbf{83.31}} 
& \underline{91.85} & 58.11 & \multicolumn{1}{c|}{71.18}  
& 68.35 & 94.72 & \multicolumn{1}{c|}{79.40} 
& 80.50  
\\
DLinear      & \multicolumn{1}{c|}{(2023)}  
& 75.91 & 84.02 & \multicolumn{1}{c|}{79.76} 
& 89.68 & 75.31  & \multicolumn{1}{c|}{81.87} 
& 89.87 & 53.79  & \multicolumn{1}{c|}{67.30 }  
& 98.65  & 94.70  & \multicolumn{1}{c|}{96.64} 
& 83.64  
\\
TimesNet     & \multicolumn{1}{c|}{(2023)}  
& 88.07 & 80.97 & \multicolumn{1}{c|}{84.37} 
& 88.83 & 74.68 & \multicolumn{1}{c|}{81.14} 
& 89.98 & 56.02  & \multicolumn{1}{c|}{69.05 }  
& 98.46  & 95.70  & \multicolumn{1}{c|}{97.06} 
& 84.85  
\\
TiDE         & \multicolumn{1}{c|}{(2023a)} 
& 76.00 & 63.00 & \multicolumn{1}{c|}{68.91} 
& 84.00 & 60.00 & \multicolumn{1}{c|}{70.18} 
& 88.00 & 50.00 & \multicolumn{1}{c|}{64.00}  
& 93.00 & 92.00 & \multicolumn{1}{c|}{92.50} 
& 74.46  
\\
iTransformer & \multicolumn{1}{c|}{(2024)}  
& 76.13 & 84.70 & \multicolumn{1}{c|}{80.19} 
& 86.15 & 62.54 & \multicolumn{1}{c|}{72.47} 
& 90.68 & 52.78 & \multicolumn{1}{c|}{66.72}  
& 97.92 & 92.03 & \multicolumn{1}{c|}{94.88} 
& 81.38  
\\
Crossformer    & \multicolumn{1}{c|}{(2023)}  
& 71.89 & 83.41 & \multicolumn{1}{c|}{77.22} 
& \underline{90.32} & 72.74 & \multicolumn{1}{c|}{80.59} 
& 89.68 & 53.63 & \multicolumn{1}{c|}{67.12}  
& 97.49 & 88.02 & \multicolumn{1}{c|}{92.52} 
& 81.53  
\\
PatchTST       & \multicolumn{1}{c|}{(2023)}  
& 87.26 & 82.14 & \multicolumn{1}{c|}{84.62} 
& 88.34 & 70.96 & \multicolumn{1}{c|}{78.70} 
& 90.64 & 55.46 & \multicolumn{1}{c|}{68.82}  
& \underline{98.84} & 93.47 & \multicolumn{1}{c|}{96.08} 
& 82.79  
\\
GPT4TS         & \multicolumn{1}{c|}{(2023)}  
& 87.68 & 81.52 & \multicolumn{1}{c|}{84.49} 
& 82.09 & \underline{81.97} & \multicolumn{1}{c|}{82.03} 
& 90.12 & 55.70 & \multicolumn{1}{c|}{68.85}  
& 98.36 & \underline{95.85} & \multicolumn{1}{c|}{97.09} 
& 85.01  
\\
Peri-midFormer & \multicolumn{1}{c|}{(2024b)} 
& 86.97 & 81.37 & \multicolumn{1}{c|}{84.08} 
& 88.66 & 74.02 & \multicolumn{1}{c|}{80.68} 
& 90.02 & 54.03 & \multicolumn{1}{c|}{67.53}  
& 98.46 & 94.06 & \multicolumn{1}{c|}{96.21} 
& 84.03  
\\
UniTS          & \multicolumn{1}{c|}{(2024)}  
& 82.42 & \underline{84.99} & \multicolumn{1}{c|}{83.69} 
& \textbf{91.32} & 73.04 & \multicolumn{1}{c|}{81.16} 
& 90.58 & \underline{62.55} & \multicolumn{1}{c|}{\underline{74.00}}  
& 98.45 & \textbf{96.19} & \multicolumn{1}{c|}{\textbf{97.31}} 
& 85.73  
\\
Timer-XL        & \multicolumn{1}{c|}{(2025)}  
& 87.60  & 82.48 & \multicolumn{1}{c|}{84.96} 
& 88.98 & 72.00 & \multicolumn{1}{c|}{79.60} 
& 90.21 & 56.30 & \multicolumn{1}{c|}{69.33}  
& 98.61 & 91.73 & \multicolumn{1}{c|}{95.05} 
& 82.23
\\
SymTime        & \multicolumn{1}{c|}{(2025)}  
& 87.93 & 81.56 & \multicolumn{1}{c|}{84.62} 
& 89.46 & 75.31 & \multicolumn{1}{c|}{81.77} 
& 90.34 & 56.96 & \multicolumn{1}{c|}{69.87}  
& \textbf{98.89} & 95.32 & \multicolumn{1}{c|}{97.07} 
& 85.39  
\\
HGTS-Former  & \multicolumn{1}{c|}{(Ours)}  
& \textbf{89.28}     &  \textbf{85.87}     & \multicolumn{1}{c|}{\textbf{87.54}}      
&  89.81     & 76.96      & \multicolumn{1}{c|}{\underline{82.89}}      
&  \textbf{91.99}     & \textbf{70.77}     & \multicolumn{1}{c|}{\textbf{79.99}}            
&  98.79     &  95.52     & \multicolumn{1}{c|}{\underline{97.13}}      
&  \textbf{86.89}     
\\ 
\bottomrule
\end{tabular}%
}
\end{table*}

\noindent $\bullet$ \textbf{Results on Anomaly Detection.}
Rapid identification of time series anomalies is crucial, as it contributes to risk prevention and decision optimization.
To validate the capabilities of our method in time series anomaly detection, we conduct experiments on four widely used anomaly detection datasets: SMD~\cite{su2019robust} and SMAP~\cite{hundman2018detecting}, MSL~\cite{hundman2018detecting}, and PSM~\cite{abdulaal2021practical}.
We adopt a processing method consistent with Anomaly Transformer to ensure fairness and use only reconstruction error as the model's optimization objective.

As shown in the Table~\ref{tab:anomaly detection}, HGTS-Former has achieved excellent performance on time series anomaly detection tasks, surpassing previous state-of-the-art methods such as Timer-XL~\cite{liu2024timer-xl}, SymTime~\cite{wang2025mitigating}, and TimesNet~\cite{wu2022timesnet} on the MSL, SMD, and SMAP datasets, and also demonstrating superior performance on the PSM dataset. 

\begin{table*}[]
\centering
\caption{ Component analysis of our model on ETT series datasets. \emph{w/o} denotes without the corresponding component.} 
\label{tab:ablation}
\resizebox{\textwidth}{!}{%
\begin{tabular}{r|c|c|c|c|c|c|c|c|c|c|c|c}
\toprule
& \multicolumn{2}{c}{{\textbf{ETTh1}}} & \multicolumn{2}{c}{{\textbf{ETTh2}}} & \multicolumn{2}{c}{{\textbf{ETTm1}}} & \multicolumn{2}{c}{{\textbf{ETTm2}}} & \multicolumn{2}{c}{{\textbf{Avg}}} &\multicolumn{2}{c}{{\textbf{Promotion}}}    \\
\multirow{-2}{*}{{}}        &MSE      &MAE       &MSE &MAE          &MSE &MAE     &MSE &MAE         &MSE &MAE         &MSE &MAE \\ \midrule
{HGTS-Former}               &0.408    &0.419       &0.347 &0.386         &0.353 &0.374    &0.267 &0.316        &0.344&0.374        &-    &-        \\
{\emph{w/o} MHSA+RoPE}      &0.449    &0.457      &0.405  &0.427         &0.395 &0.409    &0.316 &0.350        &0.391 &0.411        &\textbf{0.047}&\textbf{0.037}      \\
{\emph{w/o} Intra\-HGA}     &0.411    &0.417      &0.357  &0.388         &0.371 &0.386    &0.287 &0.328        &0.357 &0.380        &\textbf{0.013}&\textbf{0.006}      \\
{\emph{w/o} Inter\-HGA}     &0.403    &0.416      &0.360  &0.397         &0.365 &0.381    &0.291 &0.328        &0.355 &0.381        &\textbf{0.011}&\textbf{0.007}      \\ \bottomrule

\end{tabular}%
}
\end{table*}

\subsection{Comparison on Nuclear Fusion Datasets}
\subsubsection{Task Definition}
Edge Localized Modes (ELMs) are a typical magnetohydrodynamic (MHD) instability occurring in the edge pedestal region of tokamak plasmas during High-Confinement Mode (H-mode) operation. Their physical mechanism is generally described by the Peeling-Ballooning (P-B) theory, where relaxation oscillations are triggered when the edge pressure gradient and bootstrap current density jointly exceed stability thresholds, potentially facilitated by nonlinear interactions~\cite{dominski2020identification}. Temporally, ELMs exhibit a quasi-periodic ``recovery-precursor-collapse" cycle, with the collapse phase causing rapid pedestal breakdown and energy release within microseconds to milliseconds, followed by a longer rebuilding phase. In experimental diagnostic signals, ELMs display distinct time-domain signatures: they manifest as steep, high-amplitude spikes in $D_\alpha$ radiation due to increased neutral recycling from particle ejection; they are accompanied by strong high-frequency magnetic fluctuations in magnetic probe signals; and advanced diagnostics like Beam Emission Spectroscopy (BES) reveal specific evolutionary patterns in edge turbulence structures prior to the crash~\cite{joung2024tokamak}. While these transient events with clear signal signatures effectively exhaust core impurities, the massive transient heat fluxes they generate pose severe challenges to the integrity of divertor targets. 

To facilitate supervised learning and reflect both confinement regime and ELM phenomenology, we formulate a four-class classification problem with labels L-mode, No/small ELM, Type-I ELM, and Type-III ELM. L-mode denotes low-confinement discharges without a developed H-mode pedestal. For H-mode intervals, we further distinguish ELM activity by event strength and temporal signatures: No/small ELM corresponds to either the absence of identifiable ELM crashes or the presence of weak events with limited perturbation in edge-radiation and magnetic signals; Type-I ELM refers to comparatively larger-amplitude crashes that are typically associated with stronger transient releases and are of particular relevance due to their divertor heat-load implications; and Type-III ELM represents more frequent, smaller-amplitude events whose diagnostic manifestations can be less separable from other edge activities, thereby increasing the difficulty of reliable discrimination.

\subsubsection{Benchmark Results}  
As shown in Table~\ref{tab:nuclear_fusion}, we evaluate our method on ELM classification using the EAST-ELM640 dataset and compare it with Timer-XL~\cite{liu2024timer-xl}, TimeFilter~\cite{hu2025timefilter}, CrossLinear~\cite{zhou2025crosslinear}, Timer~\cite{liu2024timer}, TimeXer~\cite{wang2024timexer}, SOFTS~\cite{han2024softs}, TimeMixer~\cite{wang2024timemixer}, iTransformer~\cite{liu2023iTransformer}, and PatchTST~\cite{nie2022time}. While our approach achieves highly competitive performance in Accuracy ($89.8$) and AUC ($96.4$), which closely follow the best-performing CrossLinear (Acc = $90.1$, AUC = $97.5$), it establishes new state-of-the-art results on key class-discriminative metrics. Specifically, our method achieves the highest Precision ($80.8$) and F1 score ($79.7$) among all evaluated baselines. Coupled with a strong Recall of $78.6$, these results demonstrate our model's superior capability in minimizing false positives and maintaining an optimal balance for ELM classification.
In addition, we present the prediction results of some methods, As can be seen in Fig.~\ref{fig:ELM_visualization}.

\begin{table*}
\centering
\caption{Experimental results on the EAST-ELM640 dataset.}
\label{tab:nuclear_fusion}
\resizebox{\textwidth}{!}{%
\begin{tabular}{l|cc|cc|cc|cc|cc|cc|cc|cc|cc|cc}
\toprule 
\multirow{2}{*}{Metrics}
& \multicolumn{2}{c|}{\textbf{HGTS-Former}} 
& \multicolumn{2}{c|}{\textbf{Timer-XL}} 
& \multicolumn{2}{c|}{\textbf{TimeFilter}}
& \multicolumn{2}{c|}{\textbf{CrossLinear}} 
& \multicolumn{2}{c|}{\textbf{Timer}} 
& \multicolumn{2}{c|}{\textbf{iTransformer}} 
& \multicolumn{2}{c|}{\textbf{TimeMixer}}
& \multicolumn{2}{c|}{\textbf{TimeXer}}
& \multicolumn{2}{c|}{\textbf{SOFTS}}
& \multicolumn{2}{c}{\textbf{PatchTST}} 
\\
 \multicolumn{1}{c|}{} 
& \multicolumn{2}{c|}{\underline{(Ours)}} 
& \multicolumn{2}{c|}{\underline{(2025)}} 
& \multicolumn{2}{c|}{\underline{(2025)}} 
& \multicolumn{2}{c|}{\underline{(2025)}} 
& \multicolumn{2}{c}{\underline{(2024)}} 
& \multicolumn{2}{c}{\underline{(2024)}}
& \multicolumn{2}{c}{\underline{(2024)}}
& \multicolumn{2}{c}{\underline{(2024)}}
& \multicolumn{2}{c}{\underline{(2024)}}
& \multicolumn{2}{c}{\underline{(2023)}} 
\\ \midrule 
\multicolumn{1}{c|}{Accuracy}     
& \multicolumn{2}{c|}{\underline{89.8}}    
& \multicolumn{2}{c|}{88.6}
& \multicolumn{2}{c|}{86.0}  
& \multicolumn{2}{c|}{\textbf{90.1}}
& \multicolumn{2}{c|}{82.9} 
& \multicolumn{2}{c|}{84.7}    
& \multicolumn{2}{c|}{85.5}
& \multicolumn{2}{c|}{87.4}  
& \multicolumn{2}{c|}{86.7}
& \multicolumn{2}{c}{85.8}
\\  
\multicolumn{1}{c|}{Precision} 
& \multicolumn{2}{c|}{\textbf{80.8}}    
& \multicolumn{2}{c|}{78.8}
& \multicolumn{2}{c|}{72.9}  
& \multicolumn{2}{c|}{\underline{80.2}}
& \multicolumn{2}{c|}{73.4}
& \multicolumn{2}{c|}{72.3}    
& \multicolumn{2}{c|}{80.0}
& \multicolumn{2}{c|}{76.4}  
& \multicolumn{2}{c|}{77.6}
& \multicolumn{2}{c}{75.5} 
\\   
\multicolumn{1}{c|}{Recall} 
& \multicolumn{2}{c|}{78.6}    
& \multicolumn{2}{c|}{76.8}
& \multicolumn{2}{c|}{76.6}  
& \multicolumn{2}{c|}{\underline{79.2}}
& \multicolumn{2}{c|}{64.0} 
& \multicolumn{2}{c|}{70.3}    
& \multicolumn{2}{c|}{52.7}
& \multicolumn{2}{c|}{\textbf{79.9}}  
& \multicolumn{2}{c|}{69.5}
& \multicolumn{2}{c}{72.9}                                     
\\   
\multicolumn{1}{c|}{F1} 
& \multicolumn{2}{c|}{\textbf{79.7}}    
& \multicolumn{2}{c|}{77.3}
& \multicolumn{2}{c|}{74.7}  
& \multicolumn{2}{c|}{79.2}
& \multicolumn{2}{c|}{68.0} 
& \multicolumn{2}{c|}{70.8}    
& \multicolumn{2}{c|}{56.9}
& \multicolumn{2}{c|}{\underline{77.4}}  
& \multicolumn{2}{c|}{72.7}
& \multicolumn{2}{c}{73.2}            
\\
% \cmidrule{2-12} 
\multicolumn{1}{c|}{AUC} 
& \multicolumn{2}{c|}{96.4}    
& \multicolumn{2}{c|}{\underline{96.5}}
& \multicolumn{2}{c|}{95.0}  
& \multicolumn{2}{c|}{\textbf{97.5}}
& \multicolumn{2}{c|}{92.0} 
& \multicolumn{2}{c|}{94.5}    
& \multicolumn{2}{c|}{86.9}
& \multicolumn{2}{c|}{76.4}  
& \multicolumn{2}{c|}{95.2}
& \multicolumn{2}{c}{94.9}                    
\\ 
% \midrule
\bottomrule
\end{tabular}%
}
\end{table*}

\subsection{Component Analysis}  
In order to verify the effectiveness of each component of HGTS-Former, we conducted anomaly detection experiments on four datasets to evaluate the impact of different components on the overall performance. 
The full HGTS-Former model was used as the control group, and ablation experiments were performed by removing a single component (\emph{w/o}). 
The results are shown in the Table~\ref{tab:ablation}. After removing MHSA and RoPE, the average MSE increased by 0.047 and the average MAE increased by 0.037, indicating the importance of MHSA and RoPE. 
MHSA and RoPE has the most obvious performance improvement. After removing Intra\-HGA, the average MSE increased by 0.013 and the average MAE increased by 0.006, indicating that Intra\-HGA is critical for modeling complex relationships between local variables and improves the performance of the model. 
After removing Inter\-HGA, the average MSE increased by 0.011 and the average MAE increased by 0.007, indicating that the inter-group hypergraph can model the interaction between different variable groups, proving the effectiveness of Inter\-HGA. The experiments prove the rationality of the HGTS-Former structural design, and each component is indispensable.

\subsection{Ablation Study}

\noindent $\bullet$ \textbf{Results of Different Lookback Length.} 
We follow the TSF scaling law~\cite{shi2024scaling} to verify the impact of different input lengths on the results. 
As can be seen from Fig.~\ref{fig:ablationStudy}(a), the model performance does not improve with increasing Lookback length. This may be because to ensure the effectiveness of Intra-HyperGraph construction, a short patch length is used when the length is short, resulting in incomplete temporal patterns and thus a decrease in performance.

\noindent $\bullet$ \textbf{Results of Different Parameter $\alpha$.}  
The purpose of the hyperparameter $\alpha$ is to reduce the influence of non-hyperedge nodes on the aggregation process when performing numerical masking. As shown in Fig.~\ref{fig:ablationStudy}(b), different $\alpha$ does not have much impact on the performance of HGTS-Former, which shows that our model can well capture the potential temporal patterns within the variables and the dynamic relationship between variables.

\noindent $\bullet$ \textbf{Results of Different Backbone Blocks.} 
We verify the impact of the number of block layers on the experimental results on the ETTh1 dataset. As we can see from Fig.~\ref{fig:ablationStudy}(c), the experimental results are getting better with the increase of the number of layers, which shows that the potential of our model has not been fully developed. It also further proves the effectiveness of our proposed method of using hierarchical hypergraphs to capture the potential temporal patterns within variables and the dynamic correlations between variables.

\noindent $\bullet$ \textbf{Efficiency Analysis \& Model Parameters.} 
We analyzed the efficiency of HGTS-Former, Timer-XL, and Timer during the training phase on the ETTh1 dataset. 
For a fair comparison, the model settings are the same, where batch size is 4, layer is 1, d\_model is 512, and d\_ff is 2048. 
As shown in the Table~\ref{tab:efficiency}, although our model has the highest number of parameters, its speed and memory usage are the fastest and the least. 
% This is because the Intra-/Inter-HGA module adds additional parameters, but the attention calculation method is different from Timer-XL and Timer.***

\begin{table}[]
\centering
\caption{Efficiency analysis of HGTS-Former.}
\label{tab:efficiency}
\begin{tabular}{c|ccc}
\toprule
\textbf{Metric}      & \textbf{Parameter}    & \textbf{Memory Usage}   & \textbf{Speed}    \\ 
\midrule HGTS-Former & 10.38M    &738M   & 0.0133s/iter   \\
\midrule Timer-XL    & 3.252M    &814M    &0.0512s/iter \\
\midrule Timer       &3.252M     &830M    &0.0457s/iter \\

\bottomrule
\end{tabular}%
\end{table}

\subsection{Visualization} 
As shown in Fig.~\ref{fig:imputationResult}, we visualize the imputation results of ETTh1 and ETTh2 at different mask rates. We can find that our method achieves SOTA performance on imputation tasks, thanks to the effective capture of temporal patterns within variables and relationships between variables by HGTS-Former. 
As shown in Fig.~\ref{fig:ComparTimerXL}, we provide the prediction cases of our proposed HGTS-Former and Timer-XL~\cite{liu2024timer-xl} on 672-pred-96 and 672-pred-192. We can see that the performance of our model is better than that of Timer-XL~\cite{liu2024timer-xl}, as our model is better at capturing the temporal patterns when facing extreme changes. 
As shown in Fig.~\ref{fig:hierarchicalhypergraph}, we show the confidence matrix during the hierarchical hypergraph construction process and the completed hypergraph. HGTS-Former can effectively aggregate similar temporal patterns within variables and capture the dynamic correlation between different variables.

\subsection{Limitation Analysis}  
While HGTS-Former demonstrates strong generalization across various time series tasks, its reliance on a purely temporal modeling paradigm inherently limits its representational power.
By directly tokenizing raw time series, the model is forced to learn complex temporal dynamics entirely from scratch. 
Transforming time series into visual representations explicitly translates hidden periodicities, frequency shifts, and structural anomalies into distinct spatial textures.
Because time series and visual modalities share inherent continuity and structural similarities, this transformation enables the extraction of hierarchical features that intuitively reveal latent temporal relationships. 
Relying exclusively on temporal domain mechanisms bypasses the opportunity to leverage robust spatial pattern-recognition priors, which are highly effective in encoding these temporal hierarchies. 
Therefore, extending our hypergraph framework to incorporate visual modalities and exploit these cross-modal alignments remains a critical direction for overcoming the current information bottleneck.

\section{Conclusion and Future Works} \label{sec::conclusion}
In this paper, we propose HGTS-Former, a novel patch-based hierarchical hypergraph Transformer designed for  multivariate time series analysis. 
By constructing a hierarchical hypergraph, our model adaptively aggregates latent temporal patterns within individual variables and captures dynamic, group-wise correlations across different variables. 
Extensive experiments demonstrate that HGTS-Former achieves state-of-the-art performance across a diverse suite of tasks, including long- and short-term forecasting, missing value imputation, anomaly detection, and complex classification. 
Moving forward, to address the representational bottlenecks of purely time-domain modeling identified in our limitation analysis, our future work will explore integrating visual modalities into our framework. By incorporating visual representations, we plan to extract richer hierarchical features that intuitively reveal latent temporal relationships, fundamentally enhancing both the representation capacity and the interpretability of our time series model.

% \section*{Acknowledgment} 

\small{ 
\bibliographystyle{IEEEtran}
\bibliography{reference}
}

% that's all folks
\end{document}